\documentclass[10pt,journal,compsoc]{IEEEtran}
\ifCLASSOPTIONcompsoc
  \usepackage[nocompress]{cite}
\else
  \usepackage{cite}
\fi

\usepackage{amsfonts}
\usepackage{graphicx}
\usepackage{amsmath}
\usepackage{subfigure}
\newtheorem{definition}{Definition}
\newtheorem{problem}{Problem}
\newcommand{\eat}[1]{}
\newcommand{\myhline}{\noalign{\global\arrayrulewidth0.04cm}\hline
                      \noalign{\global\arrayrulewidth0.04pt}}
\usepackage{multirow}
\usepackage{amssymb}
\begin{document}
\title{A Comprehensive Survey of Graph Embedding: Problems, Techniques and Applications}

\author{Hongyun~Cai,
        Vincent~W.~Zheng,
        and~Kevin~Chen-Chuan~Chang% <-this % stops a space
\IEEEcompsocitemizethanks{
\IEEEcompsocthanksitem H. Cai is with Advanced Digital Sciences Center, Singapore. E-mail: hongyun.c@adsc.com.sg.
\IEEEcompsocthanksitem V. Zheng is with Advanced Digital Sciences Center, Singapore. Email: vincent.zheng@adsc.com.sg. 
\IEEEcompsocthanksitem K. Chang is with University of Illinois at Urbana-Champaign, USA. Email: kcchang@illinois.edu.
}
}

% The paper headers
\markboth{IEEE Transactions on Knowledge and Data Engineering,~Vol.~xx, No.~xx, Sept~2017}
{Cai \MakeLowercase{\textit{et al.}}: A Comprehensive Survey of Graph Embedding: Problems, Techniques and Applications}

\IEEEtitleabstractindextext{%
\begin{abstract}
Graph is an important data representation which appears in a wide diversity of real-world scenarios. Effective graph analytics provides users a deeper understanding of what is behind the data, and thus can benefit a lot of useful applications such as node classification, node recommendation, link prediction, etc. However, most graph analytics methods suffer the high computation and space cost. Graph embedding is an effective yet efficient way to solve the graph analytics problem. It converts the graph data into a low dimensional space in which the graph structural information and graph properties are maximumly preserved. In this survey, we conduct a comprehensive review of the literature in graph embedding. We first introduce the formal definition of graph embedding as well as the related concepts. After that, we propose two taxonomies of graph embedding which correspond to what challenges exist in different graph embedding problem settings and how the existing work address these challenges in their solutions. Finally, we summarize the applications that graph embedding enables and suggest four promising future research directions in terms of computation efficiency, problem settings, techniques and application scenarios.
\end{abstract}

% Note that keywords are not normally used for peerreview papers.
}

% make the title area
\maketitle

\IEEEdisplaynontitleabstractindextext

% For peerreview papers, this IEEEtran command inserts a page break and
% creates the second title. It will be ignored for other modes.
\IEEEpeerreviewmaketitle

\IEEEraisesectionheading{\section{Introduction}\label{sec:introduction}}
\IEEEPARstart{G}{raphs} naturally exist in a wide diversity of real-world scenarios, e.g., social graph/diffusion graph in social media networks, citation graph in research areas, user interest graph in electronic commerce area, knowledge graph etc. Analysing these graphs provides insights into how to make good use of the information hidden in graphs, and thus has received significant attention in the last few decades. Effective graph analytics can benefit a lot of applications, such as node classification \cite{DBLP:conf/aaai/WangCWP0Y17}, node clustering \cite{DBLP:conf/aaai/NieZL17}, node retrieval/recommendation \cite{DBLP:conf/aaai/ZhouLLLG17}, link prediction \cite{Wei:2017:CVL:3038912.3052575}, etc. For example, by analysing the graph constructed based on user interactions in a social network (e.g., retweet/comment/follow in Twitter), we can classify users, detect communities, recommend friends, and predict whether an interaction will happen between two users.

Although graph analytics is practical and essential, most existing graph analytics methods suffer the high computation and space cost. A lot of research efforts have been devoted to conducting the expensive graph analytics efficiently. Examples include the distributed graph data processing framework (e.g., GraphX \cite{DBLP:conf/osdi/GonzalezXDCFS14}, GraphLab \cite{Low:2012:DGF:2212351.2212354}), new space-efficient graph storage which accelerate the I/O and computation cost \cite{gstore}, and so on.  

\begin{figure*}[t]
  \centering
\begin{tabular}[t]{l} 
    \subfigure[Input Graph $G_1$]{\label{fig:ig} 
        \includegraphics[width=0.16\linewidth]{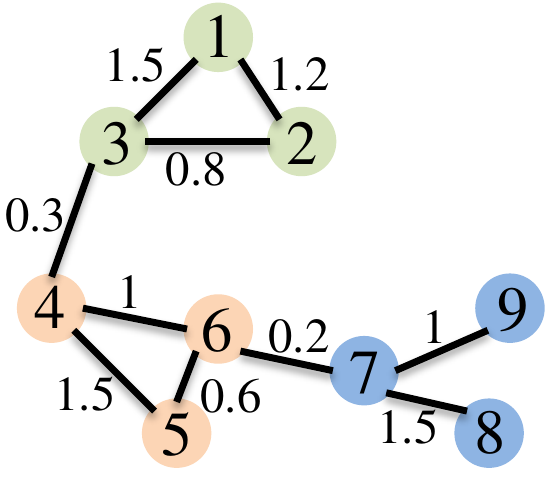}}
    \subfigure[Node Embedding]{\label{fig:ne}
        \includegraphics[width=0.195\linewidth, height=3.15cm]{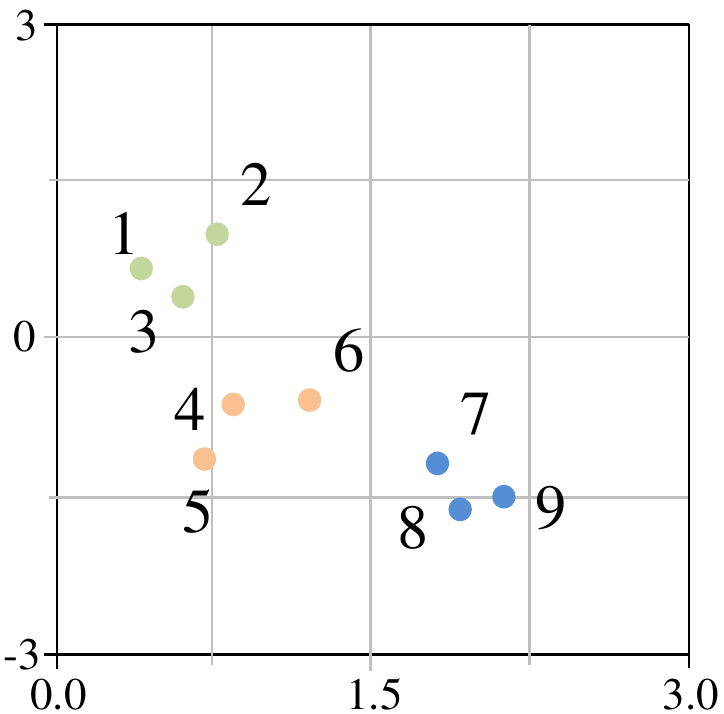}}
    \subfigure[Edge Embedding]{\label{fig:ee}
        \includegraphics[width=0.195\linewidth, height=3.15cm]{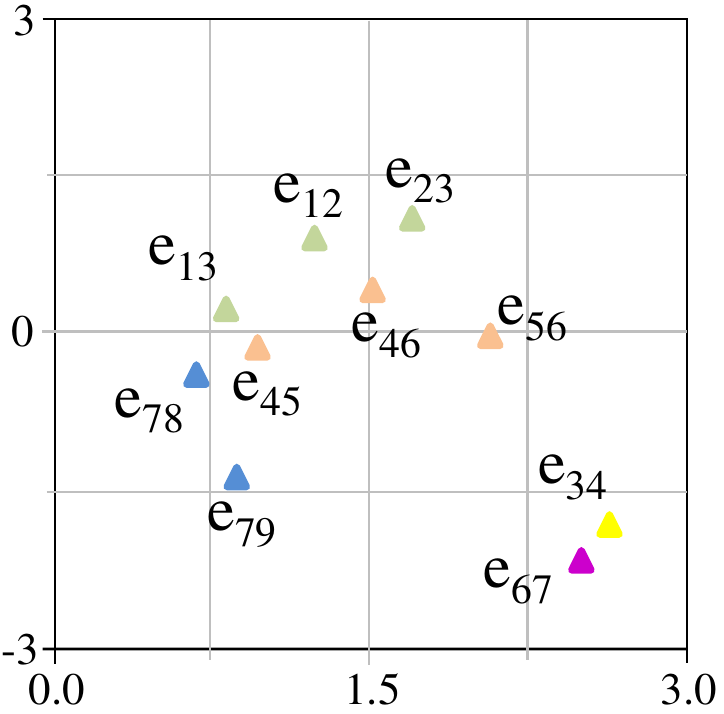}}
    \subfigure[Substructure Embedding]{\label{fig:se}
        \includegraphics[width=0.2\linewidth, height=3.15cm]{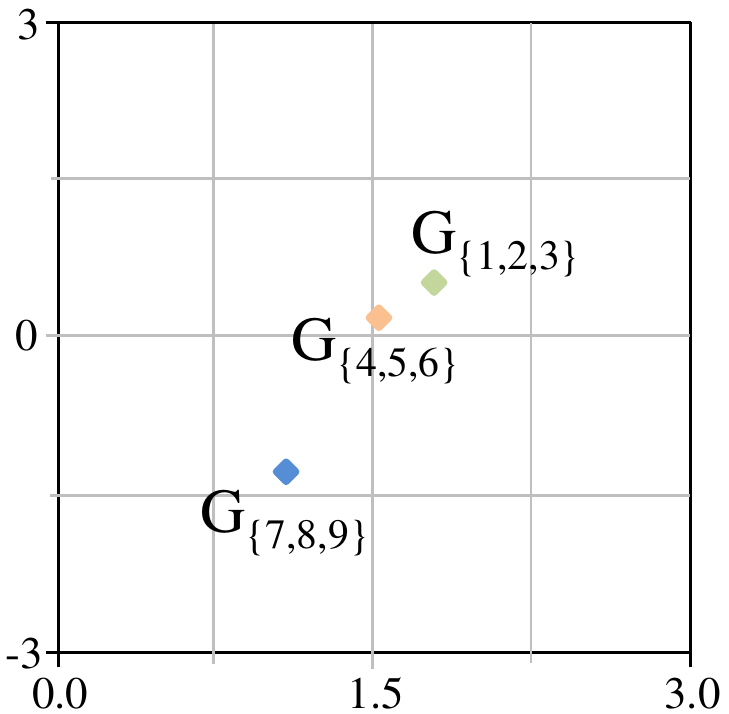}}
    \subfigure[Whole-Graph Embedding]{\label{fig:we}
        \includegraphics[width=0.2\linewidth, height=3.15cm]{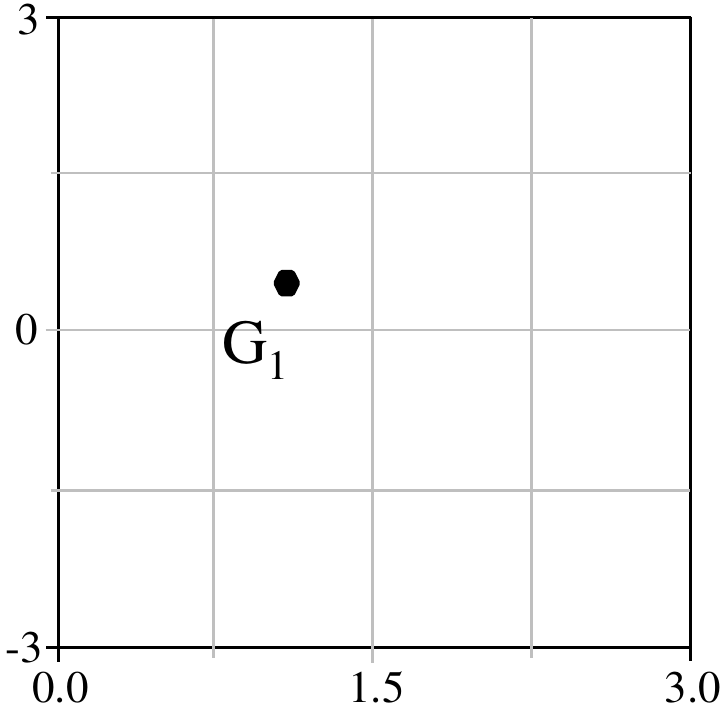}}
\end{tabular}
\vspace{-3mm}
  \caption{A toy example of embedding a graph into 2D space with different granularities. $G_{\{1,2,3\}}$ denotes the substructure containing node $v_1$, $v_2$, $v_3$.}
  \label{fig:geexample}
  %\vspace{-3mm}
\end{figure*}

In addition to the above strategies, graph embedding provides an effective yet efficient way to solve the graph analytics problem. Specifically, graph embedding converts a graph into a low dimensional space in which the graph information is preserved. By representing a graph as a (or a set of) low dimensional vector(s), graph algorithms can then be computed efficiently. There are different types of graphs (e.g., homogeneous graph, heterogeneous graph, attribute graph, etc), so the input of graph embedding varies in different scenarios. The output of graph embedding is a low-dimensional vector representing a part of the graph (or a whole graph). Fig. \ref{fig:geexample} shows a toy example of embedding a graph into a 2D space in different granularities. I.e., according to different needs, we may represent a node/edge/substructure/whole-graph as a low-dimensional vector.  More details about different types of graph embedding input and output are provided in Sec. \ref{sec:ps}.

In the early 2000s, graph embedding algorithms were mainly designed to reduce the high dimensionality of the non-relational data by assuming the data lie in a low dimensional manifold. Given a set of non-relational high-dimensional data features, a similarity graph is constructed based on the pairwise feature similarity. Then, each node in the graph is embedded into a low-dimensional space where connected nodes are closer to each other. Examples of this line of researches are introduced in Sec. \ref{sec:mf}. Since 2010, with the proliferation of graph in various fields, research in graph embedding started to take a graph as the input and leverage the auxiliary information (if any) to facilitate the embedding. On the one hand, some of them focus on representing a part of the graph (e.g., node, edge, substructure) (Figs. \ref{fig:ne}-\ref{fig:se}) as one vector. To obtain such embedding, they either adopt the state-of-the-art deep learning techniques (Sec. \ref{sec:dl}) or design an objective function to optimize the edge reconstruction probability (Sec. \ref{sec:ml}). On the other hand, there is also some work concentrating on embedding the whole graph as one vector (Fig. \ref{fig:we}) for graph level applications. Graph kernels (Sec. \ref{sec:gk}) are usually designed to meet this need.
\eat{\begin{figure}[t]
  \centering
    \includegraphics[width=0.5\linewidth]{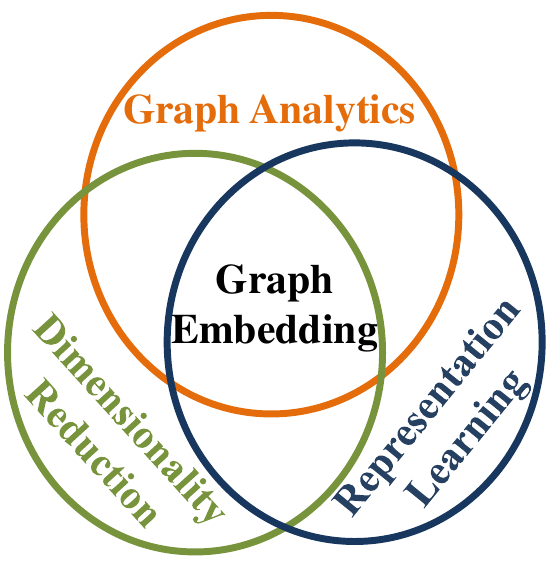}
  \caption{Relations To Other Problems}
  \label{fig:pb}
\end{figure}}

The problem of graph embedding is related to two traditional research problems, i.e., graph analytics \cite{DBLP:conf/sigmod/SatishSPSPHSYD14} and representation learning \cite{DBLP:journals/pami/BengioCV13}. Particularly, graph embedding aims to \textbf{represent} a \textbf{graph} as \textbf{low dimensional} vectors while the graph structures are preserved. On the one hand, graph analytics aims to mine useful information from graph data. On the other hand, representation learning obtains data representations that make it easier to extract useful information when building classifiers or other predictors \cite{DBLP:journals/pami/BengioCV13}. Graph embedding lies in the overlap of the two problems and focuses on learning the low-dimensional representations. 
Note that we distinguish graph representation learning and graph embedding in this survey. Graph representation learning does not require the learned representations to be low dimensional. For example, \cite{DBLP:journals/tkde/MahmoodSAR17} represents each node as a vector with dimensionality equals to the number of nodes in the input graph. Every dimension denotes the geodesic distance of a node to each other node in the graph. 

Embedding graphs into low dimensional spaces is not a trivial task. The challenges of graph embedding depend on the \textbf{problem setting}, which consists of embedding input and embedding output. In this survey, we divide the \textbf{input graph} into four categories, including \textit{homogeneous graph, heterogeneous graph, graph with auxiliary information} and \textit{graph constructed from non-relational data}. Different types of embedding input carry different information to be preserved in the embedded space and thus pose different challenges to the problem of graph embedding. For example, when embedding a graph with structural information only, the connections between nodes are the target to be preserved. However, for a graph with node label or attribute information, the auxiliary information provides graph property from other perspectives, and thus may also be considered during the embedding. 
Unlike embedding input which is given and fixed, the \textbf{embedding output} is task driven. For example, the most common type of embedding output is node embedding which represents close nodes as similar vectors. Node embedding can benefit node related tasks such as node classification, node clustering, etc. However, in some cases, the tasks may be related to higher granularity of a graph e.g., node pairs, subgraph, whole graph. Hence, the first challenge in terms of embedding output is to find a suitable embedding output type for the application of interest. We categorize four types of graph embedding output, including \textit{node embedding, edge embedding, hybrid embedding} and \textit{whole-graph embedding}. Different output granularities have different criteria for a ``good'' embedding and face different challenges. For example, a good \textit{node embedding} preserves the similarity to its neighbouring nodes in the embedded space. In contrast, a good \textit{whole-graph embedding} represents a whole graph as a vector so that the graph-level similarity is preserved.

In observations of the challenges faced in different problem settings, we propose two taxonomies of graph embedding work, by categorizing graph embedding literature based on the problem settings and the embedding techniques. These two taxonomies correspond to what challenges exist in graph embedding and how existing studies address these challenges. In particular, we first introduce different settings of graph embedding problem as well as the challenges faced in each setting. Then we describe how existing studies address these challenges in their work, including their insights and their technical solutions. 

Note that although a few attempts have been made to survey graph embedding (\cite{gesurvey, dlsurvey, DBLP:journals/tkde/WangMWG17}), they have the following two limitations. First, they usually propose only one taxonomy of graph embedding techniques. None of them analyzed graph embedding work from the perspective of problem setting, nor did they summarize the challenges in each setting. 
Second, only a limited number of related work are covered in existing graph embedding surveys. E.g., \cite{gesurvey} mainly introduces twelve representative graph embedding algorithms, and \cite{DBLP:journals/tkde/WangMWG17} focuses on knowledge graph embedding only. Moreover, there is no analysis on the insight behind each graph embedding technique. A comprehensive review of existing graph embedding work and a high level abstraction of the insight for each embedding technique can foster the future researches in the field.

\subsection{Our Contributions}
Below, we summarize our major contributions in this survey.$\!$ 

\vspace{0.04in}\noindent
$\bullet$ We propose a taxonomy of graph embedding based on problem settings and summarize the challenges faced in each setting. We are the first to categorize graph embedding work based on problem setting, which brings new perspectives to understanding existing work.

\vspace{0.04in}\noindent
$\bullet$ We provide a detailed analysis of graph embedding techniques. Compared to existing graph embedding surveys, we not only investigate a more comprehensive set of graph embedding work, but also present a summary of the insights behind each technique. In contrast to simply listing how the graph embedding was solved in the past, the summarized insights answer the questions of why the graph embedding can be solved in a certain way. This can serve as an insightful guideline for future research. 

\vspace{0.04in}\noindent
$\bullet$ We systematically categorize the applications that graph embedding enables and divide the applications as node related, edge related and graph related. For each category, we present detailed application scenarios as the reference.

\vspace{0.04in}\noindent
$\bullet$ We suggest four promising future research directions in the field of graph embedding in terms of computational efficiency, problem settings, solution techniques and applications. For each direction, we provide a thorough analysis of its disadvantages (deficiency) in current work and propose future research direction(s).

\subsection{Organization of The Survey}
\begin{figure}[t]
  \centering
    \includegraphics[width=\linewidth]{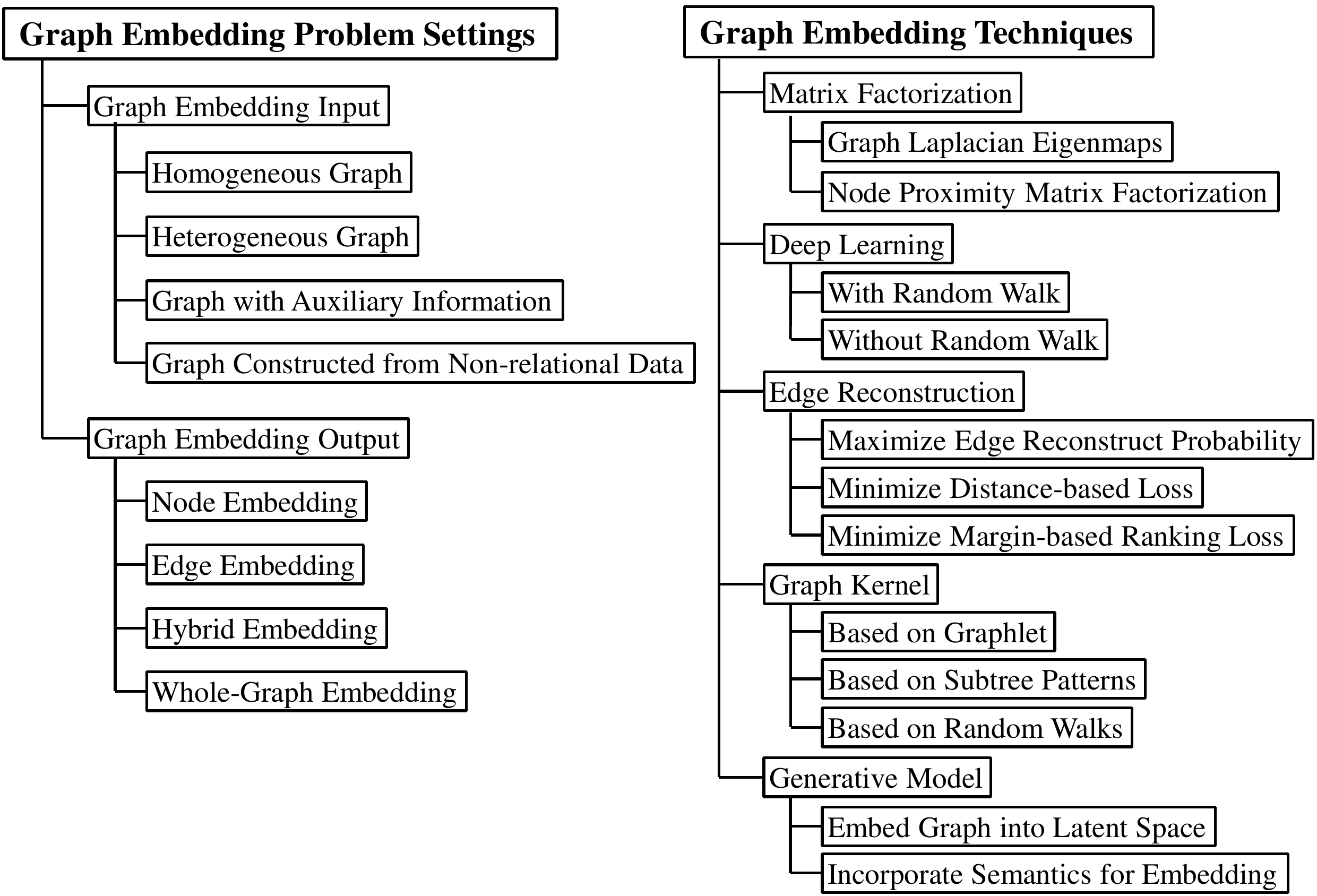}
    \vspace{-5mm}
  \caption{Graph embedding taxonomies by problems and techniques.}
  \label{fig:taxonomy}
  %\vspace{-2mm}
\end{figure}

The rest of this survey is organized as follows. In Sec. \ref{sec:pf}, we introduce the definitions of the basic concepts required to understand the graph embedding problem, and then provide a formal problem definition of graph embedding. In the next two sections, we provide two taxonomies of graph embedding, where the taxonomy structures are illustrated in Fig. \ref{fig:taxonomy}. Sec. \ref{sec:ps} compares the related work based on the problem settings and summarizes the challenges faced in each setting. In Sec. \ref{sec:get}, we categorize the literature based on the embedding techniques.The insights behind each technique are abstracted, and a detailed comparison of different techniques is provided at the end. After that, we present the applications that graph embedding enables in Sec. \ref{sec:apps}. We then discuss four potential future research directions in Sec. \ref{sec:fd} and concludes this survey in Sec. \ref{sec:conc}. 

\section{Problem Formalization}
\label{sec:pf}
In this section, we first introduce the definition of the basic concepts in graph embedding, and then provide a formal definition of the graph embedding problem. 

\subsection{Notation and Definition}
\label{sec:nd}
The detailed descriptions of the notations used in this survey can be found in Table \ref{tab:notations}.

\begin{table}[t]
\caption{Notations used in this paper.}
\vspace{-3mm}
\label{tab:notations}
\centering
\tabcolsep=0.05cm
\scriptsize
{\renewcommand{\arraystretch}{1.2}
\begin{tabular} {  l|l p{7cm} } \hline
\textbf{Notations}& \textbf{Descriptions} \\ \hline
$|\cdot|$ & The cardinality of a set \\ \hline
$\mathcal{G}$ = $(V,E)$& Graph $\mathcal{G}$ with nodes set $V$ and edges set $E$\\ \hline
$\hat{\mathcal{G}}$ = $(\hat{V}, \hat{E})$ & A substructure of graph $\mathcal{G}$, where $\hat V \subseteq V,\hat E \subseteq E$ \\  \hline
$v_i$, $e_{ij}$& A node $v_i \in V$ and an edge $e_{ij} \in E$ connecting $v_i$ and $v_j$  \\ \hline
$A$& The adjacent matrix of $\mathcal{G}$ \\ \hline
$A_{i}$ & The $i$-th row vector of matrix $A$ \\ \hline
$A_{i,j}$ & The $i$-th row and $j$-th column in matrix $A$ \\ \hline
$f_v(v_i)$, $f_e(e_{ij})$& Type of node $v_i$ and type of edge $e_{ij}$ \\\hline
$\mathcal{T}^v$,  $\mathcal{T}^e$& The node type set and edge type set\\ \hline
$\mathcal{N}_k(v_i)$ & The k nearest neighbours of node $v_i$\\\hline
$X \in \mathbb{R}^{|V| \times N}$& A feature matrix, each row $X_i$ is a $N$-dimensional vector for $v_i$  \\\hline
$y_i$, $y_{ij}$, $y_{\hat{\mathcal{G}}}$  & The embedding of node $v_i$, edge $e_{ij}$, and structure $\hat{\mathcal{G}}$  \\\hline
$d$ & The dimensionality of the embedding \\\hline
\multirow{2}{*}{$<h,r,t>$} & A knowledge graph triplet, with head entity $h$, \\
& tail entity $t$ and the relation between them $r$ \\ \hline
$s_{ij}^{(1)}$, $s_{ij}^{(2)}$ & First- and second-order proximity between node $v_i$ and $v_j$\\\hline
$c$ & An information cascade\\\hline
$\mathcal{G}^c = (V^c, E^c)$ & A cascade graph which adopts the cascade $c$ \\\hline
\end{tabular}}
\end{table}

\begin{definition}
\label{def:graph}
A \textbf{graph} is $\mathcal{G}$ = $(V,E)$, where $v \in V$ is a node and $e \in E$ is an edge. $\mathcal{G}$ is associated with a node type mapping function $f_v:V \to \mathcal{T}^v$ and an edge type mapping function $f_e:E \to \mathcal{T}^e$. 
\end{definition}
$\mathcal{T}^v$ and $\mathcal{T}^e$ denote the set of node types and edge types, respectively. Each node $v_i \in V$ belongs to one particular type, i.e., $f_v(v_i) \in \mathcal{T}^v$. Similarly, for $e_{ij} \in E$, $f_e(e_{ij}) \in \mathcal{T}^e$.

\begin{definition}
\label{def:homograph}
A \textbf{homogeneous graph} $\mathcal{G}_{homo}$ = $(V,E)$ is a graph in which $|\mathcal{T}^v| = |\mathcal{T}^e| = 1$. All nodes in $\mathcal{G}$ belong to a single type and all edges belong to one single type.
\end{definition}

\begin{definition}
\label{def:hetegraph}
A \textbf{heterogeneous graph}  $\mathcal{G}_{hete}$ = $(V,E)$ is a graph in which $|\mathcal{T}^v| >1$ and/or $|\mathcal{T}^e| > 1$. 
\end{definition}

\begin{figure}[t]
  \centering
    \includegraphics[width=0.6\linewidth]{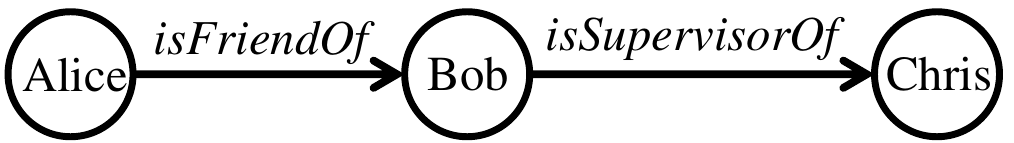}
    \vspace{-2mm}
  \caption{A toy example of knowledge graph.}
  \label{fig:kgexample}
  \vspace{-2mm}
\end{figure}

\begin{definition}
\label{def:knowgraph}
A \textbf{knowledge graph} $\mathcal{G}_{know}$ = $(V,E)$ is a directed graph whose nodes are \textit{entities} and edges are \textit{subject-property-object} triple facts. Each edge of the form (\textit{head entity}, \textit{relation}, \textit{tail entity}) (denoted as $<h,r,t>$)  indicates a relationship of $r$ from entity $h$ to entity $t$. 
\end{definition}
$h,t \in V$ are entities and $r \in E$ is the relation. In this survey, we call $<h,r,t>$ a knowledge graph triplet. For example, in Fig. \ref{fig:kgexample}, there are two triplets: $<Alice, isFriendOf, Bob>$ and $<Bob, isSupervisorOf, Chris>$. Note that the entities and relations in a knowledge graph are usually of different types \cite{DBLP:conf/aaai/WangZFC14,DBLP:conf/aaai/LinLSLZ15}. Hence, knowledge graph can be viewed as an instance of the heterogeneous graph.

The following proximity measures are usually adopted to quantify the graph property to be preserved in the embedded space.
The first-order proximity is the local pairwise similarity between only the nodes connected by edges. It compares the direct connection strength between a node pair. Formally,

\begin{definition}
\label{def:fpro}
The \textbf{first-order proximity} between node $v_i$ and node $v_j$ is the weight of the edge $e_{ij}$, i.e., $A_{i,j}$.
\end{definition}

Two nodes are more similar if they are connected by an edge with larger weight. Denote the first-order proximity between node $v_i$ and $v_j$ as $s_{ij}^{(1)}$, we have $s_{ij}^{(1)} = A_{i,j}$. Let $s_{i}^{(1)} = [s_{i1}^{(1)}, s_{i2}^{(1)}, \cdots, s_{i|V|}^{(1)}]$ denote the first-order proximity between $v_i$ and other nodes. 
Take the graph in Fig. \ref{fig:ig} as an example, the first order between $v_1$ and $v_2$ is the weight of edge $e_{12}$, denoted as $s_{12}^{(1)}=1.2$. And $s_1^{(1)}$ records the weight of edges connecting $v_1$ and other nodes in the graph, i.e., $s_1^{(1)}=[0,1.2,1.5,0,0,0,0,0,0]$.

The second-order proximity compares the similarity of the nodes' neighbourhood structures. The more similar two nodes' neighbourhoods are, the larger the second-order proximity value between them. Formally, 

\begin{definition}
\label{def:spro}
The \textbf{second-order proximity} $s_{ij}^{(2)}$ between node $v_i$ and $v_j$ is a similarity between $v_i$'s neighbourhood $s_i^{(1)}$ and $v_j$'s neighborhood $s_j^{(1)}$.
\end{definition}

Again, take Fig. \ref{fig:ig} as an example: $s_{12}^{(2)}$ is the similarity between $s_1^{(1)}$ and $s_2^{(1)}$. As introduced before, $s_1^{(1)}=[0,1.2,1.5,0,0,0,0,0,0]$ and $s_2^{(1)}=[1.2,0,0.8,0,0,$ $0,0,0,0]$. Let us consider cosine similarities $s_{12}^{(2)} = cosine(s_1^{(1)},s_2^{(1)})=0.43$ and $s_{15}^{(2)} = cosine(s_1^{(1)},s_5^{(1)})=0$. We can see that the second-order proximity between $v_1$ and $v_5$ equals to zero as $v_1$ and $v_5$ do not share any common $1$-hop neighbour. $v_1$ and $v_2$ share a common neighbour $v_3$, hence their second-order proximity $s_{12}^{(2)}$ is larger than zero.

The \textbf{higher-order proximity} can be defined likewise. For example, the $k$-th-order proximity between node $v_i$ and $v_j$ is the similarity between $s_i^{(k-1)}$ and $s_j^{(k-1)}$. Note that sometimes the higher-order proximities are also defined using some other metrics, e.g., Katz Index, Rooted PageRank, Adamic Adar, etc \cite{gesurvey}. 

It is worth noting that, in some work, the first-order and second-order proximities are empirically calculated based on the joint probability and conditional probability of two nodes. More details are discussed in Sect. \ref{sec:mdbl}.  

\begin{problem}
\label{pro:ge}
\textbf{Graph embedding:} Given the input of a graph $\mathcal{G}$ = $(V,E)$, and a predefined dimensionality of the embedding $d$ ($d \ll |V|$), the problem of graph embedding is to convert $\mathcal{G}$ into a $d$-dimensional space, in which the graph property is preserved as much as possible. The graph property can be quantified using proximity measures such as the first- and higher-order proximity. Each graph is represented as either a $d$-dimensional vector (for a whole graph) or a set of $d$-dimensional vectors with each vector representing the embedding of part of the graph (e.g., node, edge, substructure). 
\end{problem}

Fig. \ref{fig:geexample} shows a toy example of graph embedding with $d=2$. Given an input graph (Fig. \ref{fig:ig}), the graph embedding algorithms are applied to convert a node (Fig. \ref{fig:ne})/ edge (Fig. \ref{fig:ee}), substructure (Fig. \ref{fig:se})/ whole-graph (Fig. \ref{fig:we}) as a 2D vector (i.e., a point in a 2D space).
In the next two sections, we provide two taxonomies of graph embedding, by categorizing the graph embedding literature based on problem settings and embedding techniques respectively. 

\section{Problem Settings of Graph Embedding}
\label{sec:ps}
In this section, we compare existing graph embedding work from the perspective of problem setting, which consists of the embedding input and the embedding output. For each setting, we first introduce different types of graph embedding input or output, and then summarize the challenges faced in each setting at the end. 

We start with graph embedding input. As a graph embedding setting consists of both input and output, we use node embedding as an example embedding output setting during the introduction of different types of input. The reason is that although there exist various types of embedding output, the majority of graph embedding studies focus on node embedding, i.e., embedding nodes to a low dimensional space where the node similarity in the input graph is preserved. More details about node embedding and other types of embedding output are presented in Sec. \ref{sec:geo}.

\subsection{Graph Embedding Input}
\label{sec:gei}
The input of graph embedding is a graph. In this survey, we divide graph embedding input into four categories: homogeneous graph, heterogeneous graph, graph with auxiliary information and constructed graph. Each type of graph poses different challenges to graph embedding. Next, we introduce these four types of input graphs and summarize the challenges faced in each input setting.

\subsubsection{Homogeneous Graph}
\label{sec:hog}
\begin{figure}[t] 
\begin{tabular}[t]{l} 
    \subfigure[A weighted graph.]{\label{fig:weightedgraph} 
        \includegraphics[width=0.35\linewidth]{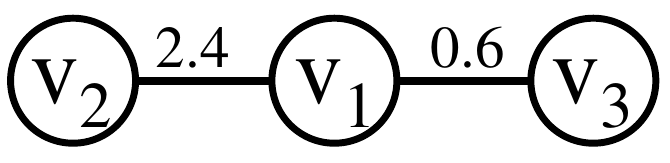}}
	\hspace{6mm}
    \subfigure[A directed graph.]{\label{fig:directedgraph}
        \includegraphics[width=0.35\linewidth]{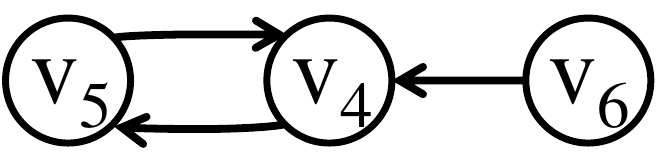}}
\end{tabular}
\vspace{-2mm}
\caption{Examples of weighted and directed graphs.} \label{fig:examplegraph}
\vspace{-2mm}
\end{figure}
The first category of input graph is the homogeneous graph (Def. \ref{def:homograph}), in which both nodes and edges belong to a single type respectively. The homogeneous graph can be further categorized as the weighted (or directed) and unweighted (or undirected) graph as the example shown in Fig. \ref{fig:examplegraph}.

\textbf{Undirected and unweighted} homogeneous graph is the most basic graph embedding input setting. A number of studies work under this setting, e.g., \cite{DBLP:conf/aaai/WangCWP0Y17, DBLP:journals/pvldb/ZhaoCSZZ13,deepwalk,DBLP:conf/ijcai/ManSLJC16, tpavnz}. They treat all nodes and edges equally, as only the basic structural information of the input graph is available.

Intuitively, the weights and directions of the edges provide more information about the graph, and may help represent the graph more accurately in the embedded space. For example, in Fig. \ref{fig:weightedgraph}, $v_1$ should be embedded closer to $v_2$ than $v_3$ because the weight of the edge $e_{1,2}$ is higher. Similarly, $v_4$ in Fig. \ref{fig:directedgraph} should be embedded closer to $v_5$ than $v_6$ as $v_4$ and $v_5$ are connected in both direction. The above information is lost in the unweighted and undirected graph. 
Noticing the advantages of exploiting the weight and direction property of the graph edges, the graph embedding community starts to explore the weighted and/or the directed graph. Some of them focus on only one graph property, i.e., either edge weight or edge direction. 
On the one hand, the \textbf{weighted graph} is considered in \cite{sdne, grarep,DBLP:conf/aaai/TianGCCL14,dnnflgr,Ahmed:2013:DLN:2488388.2488393,DBLP:conf/ijcnn/JinLLZZW16}. Nodes connected by higher-weighted edges are embedded closer to each other. However, their work is still limited to undirected graphs. 
On the other hand, some work distinguishes directions of the edges during the embedding process and preserve the direction information in the embedded space. One example of the \textbf{directed graph} is the social network graph, e.g, \cite{DBLP:conf/ijcai/LiuCLL16}. Each user has both followership and followeeship with other users. However, the weight information is unavailable for the social user links. 
Recently, a more general graph embedding algorithm is proposed, in which both weight and direction properties are considered. In other words, these algorithms (e.g., \cite{line,DBLP:conf/aaai/ZhouLLLG17,node2vec}) can process \textbf{both directed and undirected, as well as weighted and unweighted graph}.

\vspace{0.05in}\noindent
\textbf{Challenge:} \textit{How to capture the diversity of connectivity patterns observed in graphs?} Since only structural information is available in homogeneous graphs, the challenge of homogeneous graph embedding lies in how to preserve these connectivity patterns observed in the input graphs during embedding.
   
\eat{
[where to put this part: asymmetric: [42](following, Asymmetric Multi-Faceted Ranking)[44](Asymmetric Proximity)[43](both symmetric and asymmetric)[15](Asymmetric Transitivity)]}

\subsubsection{Heterogeneous Graph}
\label{sec:heg}
The second category of input is the heterogeneous graph (Def. \ref{def:hetegraph}), which mainly exist in the three scenarios below.

\textbf{Community-based Question Answering (cQA) sites.} cQA is an Internet-based crowdsourcing service that enables users to post questions on a website, which are then answered by other users \cite{cqahsnl}. Intuitively, there are different types of nodes in a cQA graph, e.g., question, answer, user.
Existing cQA graph embedding methods distinguish from each other in terms of the links they exploit as summarized in Table \ref{tab:cqagraph}, where $(i,j,k,o,p)$ denotes that the $j$-th answer provided by user $k$ obtains more votes (i.e., thumb-ups) than the $o$-th answer of user $p$ for question $i$.
\begin{table}[t]
\caption{Graph Embedding Algorithms for cQA sites}
\vspace{-3mm}
\label{tab:cqagraph}
\centering
\tabcolsep=0.05cm
\scriptsize
\renewcommand{\arraystretch}{1.2}
\begin{tabular} {  l|l } \hline
\textbf{GE Algorithm}& \textbf{Links Exploited} \\ \hline
\cite{DBLP:conf/ijcai/ZhaoYCHZ16}&  user-user, user-question \\ \hline
\cite{DBLP:conf/aaai/LuK17} &  user-user, user-question, question-answer \\ \hline
\cite{cqahsnl} & user-user,  question-answer, user-answer \\\hline
\cite{DBLP:conf/aaai/ZhaoLZCHZ17}  & users' asymmetric following links, a ordered tuple $(i,j,k,o,p)$ \\\hline
\end{tabular}
\vspace{-2mm}
\end{table}

\textbf{Multimedia Networks.} A multimedia network is a network containing multimedia data, e.g., image, text, etc. For example, both \cite{hneda} and \cite{DBLP:journals/tomccap/ZhangSLWC16} embed the graphs containing two types of nodes (image and text) and three types of links (the co-occurrence of image-image, text-text and image-text). \cite{DBLP:conf/iccv/GengZBC15} processes a social curation with user node and image node. It exploits user-image links to embed users and images into the same space so that they can be directly compared for image recommendation. In \cite{DBLP:journals/tip/WuLSYZRZ16}, a click graph is considered which contains images and text queries. The image-query edge indicates a click of an image given a query, where the click count serves as the edge weight. 

\textbf{Knowledge Graphs.} In a knowledge graph (Def. \ref{def:knowgraph}), the entities (nodes) and relations (edges) are usually of different types. For example, in a film related knowledge graph constructed from Freebase \cite{Bollacker:2008:FCC:1376616.1376746}, the types of entities can be ``\textit{director}'', ``\textit{actor}'', ``\textit{film}'', etc. The types of relations can be ``\textit{produce}'', ``\textit{direct}'', ``\textit{act$\_$in}''. A lot of efforts have been devoted to embeding knowledge graphs (e.g., \cite{DBLP:conf/coling/FengHYZ16,DBLP:conf/aaai/WuSYLZZ15,DBLP:conf/aaai/ShiW17}). We will introduce them in details in Sec. \ref{sec:mlml}.

\eat{Another common example is the knowledge graph \cite{DBLP:conf/coling/FengHYZ16,DBLP:conf/aaai/WuSYLZZ15,DBLP:conf/aaai/ShiW17}
\cite{cdkge,DBLP:conf/aaai/BordesWCB11,DBLP:conf/nips/BordesUGWY13,DBLP:journals/ml/BordesGWB14,DBLP:conf/acl/0005HZ16,DBLP:conf/aaai/LinLSLZ15,DBLP:conf/aaai/WangZFC14,DBLP:conf/coling/FengHYZ16,DBLP:conf/aaai/WuSYLZZ15,DBLP:conf/aaai/ShiW17}} 

Other heterogeneous graphs also exist. For instance, \cite{DBLP:conf/huc/OchiNYSARM16} and \cite{jcnss16geo} work on the mobility data graph, in which the station (s), role (r) and company (c) nodes are connected by three types of links (s-s, s-r, s-c). \cite{rlfmerri} embeds a Wikipedia graph with three types of nodes (entity (e), category (c) and word (w)) and three types of edges (e-e, e-c, w-w). In addition to the above graphs, there are some general heterogeneous graphs in which the types of nodes and edges are not specifically defined \cite{DBLP:conf/aaai/LiuZZZCWY17,DBLP:conf/icdm/GuiLTJNH16, dong2017metapath2vec}.

\vspace{0.05in}\noindent
\textbf{Challenge:} \textit{How to explore global consistency between different types of objects, and how to deal with the imbalances of objects belonging to different types, if any?} Different types of objects (e.g., nodes, edges) are embedded into the same space in heterogeneous graph embedding. How to explore the global consistency between them is a problem. Moreover, there may exist imbalance between objects of different types. This data skewness should be considered in embedding.

\subsubsection{Graph with Auxiliary Information}
\label{sec:gwai}
The third category of input graph contains auxiliary information of a node/edge/whole-graph in addition to the structural relations of nodes (i.e., $E$). Generally, there are five different types of auxiliary information as listed in Table \ref{tab:auinfo}.

\begin{table}[t]
\caption{Comparison of Different Types of Auxiliary Information in Graphs}
\vspace{-3mm}
\label{tab:auinfo}
\centering
\tabcolsep=0.05cm
\scriptsize
\renewcommand{\arraystretch}{1.2}
\begin{tabular} {  l|l } \hline
\textbf{Auxiliary Information}& \textbf{Description} \\ \hline
label&  categorical value of a node/edge, e.g., class information \\ \hline
attribute & categorical or continuous value of a node/edge, \\
& e.g., property information\\ \hline
node feature & text or image feature for a node \\\hline
information propagation  & the paths of how the information is propagated in graphs \\\hline
knowledge base & text associated with or facts between  knowledge concepts\\\hline
\end{tabular}
\vspace{-2mm}
\end{table}

\textbf{Label}: Nodes with different labels should be embedded far away from each other. In order to achieve this, \cite{DBLP:conf/acl/LiZZ16} and \cite{DBLP:conf/ijcai/TuZLS16} jointly optimize the embedding objective function together with a classifier function. \cite{journals/jmlr/ShervashidzeSLMB11} puts a penalty on the similarity between nodes with different labels. \cite{DBLP:conf/aaai/NikolentzosMV17} considers node labels and edge labels when calculating different graph kernels. \cite{DBLP:conf/acl/GuoWWWG15} and \cite{DBLP:journals/tkde/GuoWWWG17} embed a knowledge graph, in which the entity (node) has a semantic category. \cite{DBLP:conf/ijcai/XieLS16} embeds a more complicated knowledge graph with the entity categories in a hierarchical structure, e.g., the category ``book'' has two sub-categories ``author'' and ``written$\_$work''. 

\textbf{Attribute}: In contrast to a label, an attribute value can be discrete or continuous. For example, \cite{DBLP:conf/icml/DaiDS16} embeds a graph with discrete node attribute value (e.g., the atomic number in a molecule). In contrast, \cite{Wei:2017:CVL:3038912.3052575} represents the node attribute as a continuous high-dimensional vector (e.g., user attribute features in social networks). \cite{lcnnfg} deals with both discrete and continuous attributes for nodes and edges.

\textbf{Node feature}: Most node features are text, which are provided either as a feature vector for each node \cite{nrlrti,DBLP:conf/icdm/ZhangYZZ16} or as a document \cite{DBLP:conf/icdm/LeL14,DBLP:conf/aaai/0005HMZ17,DBLP:conf/aaai/YaoZWJZZC17,Wang:2016:TRL:3060621.3060801}. For the latter, the documents are further processed to extract feature vectors using techniques such as bag-of-words \cite{DBLP:conf/icdm/LeL14}, topic modelling \cite{DBLP:conf/aaai/0005HMZ17,DBLP:conf/aaai/YaoZWJZZC17}, or treating ``word'' as one type of node \cite{Wang:2016:TRL:3060621.3060801}.
Other types of node features, such as  image features \cite{hneda}, are also possible. Node features enhance the graph embedding performance by providing rich and unstructured information, which is available in many real-world graphs. Moreover, it makes inductive graph embedding possible \cite{rsslge}.   

\textbf{Information propagation}: An example of information propagation is ``retweet'' in Twitter. In \cite{DBLP:conf/www/LiMGM17}, given a data graph $\mathcal{G} = (V, E)$, a cascade graph $\mathcal{G}^c = (V^c, E^c)$ is constructed for each cascade $c$, where $V^c$ are the nodes that have adopted $c$ and $E^c$ are the edges with both ends in $V^c$. They then embed $\mathcal{G}^c$ to predict the increment of cascade size. Differently, \cite{DBLP:journals/ijmir/ZhaoZWHC16} aims to embed the users and content information, such that the similarity between their embedding indicates a diffusion probability. Topo-LSTM \cite{DBLP:conf/icdm/WangZLC17} considers a cascade as not merely a sequence of nodes, but  a dynamic directed acyclic graphs for embedding. 

\textbf{Knowledge base}: The popular knowledge bases include Wikipedia \cite{mmbeflskg}, Freebase \cite{Bollacker:2008:FCC:1376616.1376746}, YAGO \cite{DBLP:conf/www/SuchanekKW07}, DBpedia \cite{DBLP:journals/ws/BizerLKABCH09}, etc. Take Wikipedia as an example, the concepts are entities proposed by users and text is the article associated with the entity. \cite{mmbeflskg} uses knowledge base to learn a social knowledge graph from a social network by linking each social network user to a given set of knowledge concepts. \cite{Xiong:2017:ESR:3038912.3052558} represents queries and documents in the entity space (provided by a knowledge base) so that the academic search engine can understand the meaning of research concepts in queries.

Other types of auxiliary information include user check-in data (user-location) \cite{DBLP:conf/icdm/AlharbiZ16}, user item preference ranking list \cite{DBLP:conf/sigir/ZhangW16}, etc. Note that the auxiliary information is not just limited to one type. For instance, \cite{rsslge} and \cite{DBLP:journals/corr/KipfW16} consider both label and node feature information. \cite{DBLP:conf/ijcai/PanWZZW16} utilizes node contents and  labels to assist the graph embedding process.

\vspace{0.05in}\noindent
\textbf{Challenge:} \textit{How to incorporate the rich and unstructured information so that the learnt embeddings are both representing the topological structure and discriminative in terms of the auxiliary information?} The auxiliary information helps to define node similarity in addition to graph structural information. The challenges of embedding graph with auxiliary information is how to combine these two information sources to define the node similarity to be preserved.

\subsubsection{Graph Constructed from Non-relational Data} \label{sec:cg}
As the last category of input graph is not provided, but constructed from the non-relational input data by different strategies. This usually happens when the input data is assumed to lie in a low dimensional manifold.   

In most cases, the input is a feature matrix $X \in \mathbb{R}^{|V| \times N}$ where each row $X_i$ is an $N$-dimensional feature vector for the $i$-th training instance. A similarity matrix $S$ is constructed by calculating $S_{ij}$ using the similarity between ($X_i$, $X_j$).
There are usually two ways to construct a graph from $S$.
A straightforward way is to directly treat $S$ as the adjacency matrix $A$ of an invisible graph \cite{DBLP:conf/nips/HofmannB94}. However, \cite{DBLP:conf/nips/HofmannB94} is based on the Euclidean distance and it does not consider the neighbouring nodes when calculating $S_{ij}$. If $X$ lies on or near a curved manifold, the distance between $X_i$ and $X_j$ over the manifold is much larger than their Euclidean distance \cite{dlsurvey}.
To address these issues, other methods (e.g., \cite{DBLP:conf/ijcai/HanS16,DBLP:journals/pami/YinGL16,DBLP:conf/mm/TangNJ16}
\eat{\cite{DBLP:conf/aaai/NieZL17,DBLP:journals/tkde/AllabLN17,Balasubramanian7,Roweis2000,DBLP:conf/icml/WeinbergerSS04,DBLP:journals/jmlr/ShawJ07,DBLP:conf/icml/ShawJ09,DBLP:conf/nips/HeN03,DBLP:journals/tkde/ChenTTJ15,DBLP:conf/mm/CaiHH07,DBLP:conf/ijcai/HanS16,DBLP:journals/pami/YinGL16,DBLP:conf/mm/TangNJ16}}) construct a K nearest neighbour (KNN) graph from $S$ first and estimate the adjacency matrix $A$ based on the KNN graph. For example, Isomap \cite{Balasubramanian7} incorporates the geodesic distance in $A$. It first constructs a KNN graph from $S$, and then finds the shortest path between two nodes as the geodesic distance between them. To reduce the cost of KNN graph construction ($O(|V|^2)$), \cite{DBLP:journals/ijon/JiangFWHH16} constructs an Anchor graph instead, whose cost is $O(|V|)$ in terms of both time and space consumption. They first obtain a set of clustering centers as virtual anchors and find the K nearest anchors of each node for anchor graph construction.

Another way of graph construction is to establish edges between nodes based on the nodes' co-occurrence. For example, to facilitate image related applications (e.g., image segmentation, image classification), researchers (e.g., \cite{DBLP:journals/corr/MontiBMRSB16, DBLP:journals/tkde/ChenTTJ15, DBLP:conf/nips/DefferrardBV16}) construct a graph from each image by treating pixels as nodes and the spatial relations between pixels as edges. \cite{Zhang:2017:RPA:3038912.3052601} extracts three types of nodes (location, time and message) from the GTMS record and therefore forms six types of edges between these nodes. \cite{Ren:2016:LNR:2939672.2939822} generates a graph using entity mention, target type and text feature as nodes, and establishes three kinds of edges: mention-type, mention-feature and type-type.

In addition to the above pairwise similarity based and node co-occurrence based methods, other graph construction strategies have been designed for different purposes. For example, \cite{DBLP:journals/pami/YanXZZYL07} constructs an intrinsic graph to capture the intraclass compactness, and a penalty graph to characterize the interclass separability.  The former is constructed by connecting each data point with its neighbours of the same class, while the latter connects the marginal points across different classes. \cite{DBLP:conf/aaai/GongTYF14} constructs a signed graph to exploit the label information. Two nodes are connected by a positive edge if they belong to the same class, and a negative edge if they are from two classes. \cite{DBLP:conf/kdd/SunJY08} includes all instances with a common label into one hyperedge to capture their joint similarity. In \cite{Lin:2005:SML:1101149.1101193}, two feedback graphs are constructed to gather together relevant pairs and keep away irrelevant ones after embedding. In the positive graph, two nodes are connected if they are both relevant. In the negative graph, two nodes are connected only when one node is relevant and the other is irrelevant.

\vspace{0.05in}\noindent
\textbf{Challenge:} \textit{How to construct a graph that encodes the pairwise relations between instances and how to preserve the generated node proximity matrix in the embedded space?} The first challenge faced by embedding graphs constructed from non-relational data is how to compute the relations between the non-relational data and construct such a graph. After the graph is constructed, the challenge becomes the same as in other input graphs, i.e., how to preserve the node proximity of the constructed graph in the embedded space.

\subsection{Graph Embedding Output}
\label{sec:geo}
The output of graph embedding is a (set of) low dimensional vector(s) representing (part of) a graph. Based on the output granularity, we divide graph embedding output into four categories, including node embedding, edge embedding, hybrid embedding and whole-graph embedding. Different types of embedding facilitate different applications. 

Unlike embedding input which is fixed and given, the embedding output is task driven. For example, node embedding can benefit a wide variety of node related graph analysis tasks. By representing each node as a vector, the node related tasks such as node clustering, node classification, can be performed efficiently in terms of both time and space. However, graph analytics tasks are not always at node level. In some scenarios, the tasks may be related to higher granularity of a graph, such as node pairs, subgraph, or even a whole graph. Hence, the first \textbf{challenge} in terms of embedding output is \textit{how to find a suitable type of embedding output which meets the needs of the specific application task.} 

\subsubsection{Node Embedding}
\label{sec:ne}
As the most common embedding output setting, node embedding represents each node as a vector in a low dimensional space. Nodes that are ``close'' in the graph are embedded to have similar vector representations. The differences between various graph embedding methods lie in how they define the ``closeness'' between two nodes. First-order proximity (Def. \ref{def:fpro}) and second-order proximity (Def. \ref{def:spro}) are two commonly adopted metrics for pairwise node similarity calculation. In some work, higher-order proximity is also explored to certain extent. For example, \cite{grarep} captures the $k$-step ($k=1,2,3,\cdots$) neighbours relations in their embedding. Both \cite{DBLP:conf/aaai/WangCWP0Y17} and \cite{DBLP:conf/cikm/CavallariZCCC17} consider two nodes belonging to the same community as embedded closer. 

\vspace{0.05in}\noindent
\textbf{Challenge:} \textit{How to define the pairwise node proximity in various types of input graph and how to encode the proximity in the learnt embeddings?} The challenges of node embedding mainly come from defining the node proximity in the input graph. In Sec \ref{sec:gei}, we have elaborated the challenges of node embedding with different types of input graphs. 

Next, we will introduce other types of embedding output as well as the new challenges posed by these outputs.  

\subsubsection{Edge Embedding}
\label{sec:ee}
In contrast to node embedding, edge embedding aims to represent an edge as a low-dimensional vector. Edge embedding is useful in the following two scenarios.

Firstly, knowledge graph embedding (e.g., \cite{DBLP:conf/aaai/BordesWCB11,DBLP:conf/nips/BordesUGWY13,DBLP:journals/ml/BordesGWB14}
\eat{\cite{DBLP:conf/aaai/BordesWCB11,DBLP:conf/nips/BordesUGWY13,DBLP:journals/ml/BordesGWB14,DBLP:conf/acl/0005HZ16,DBLP:conf/ijcai/XieLS16,DBLP:conf/aaai/LinLSLZ15,DBLP:conf/acl/GuoWWWG15,DBLP:conf/aaai/0005HMZ17,DBLP:conf/aaai/WangZFC14,DBLP:conf/coling/FengHYZ16,DBLP:conf/aaai/WuSYLZZ15,DBLP:conf/aaai/ShiW17,Wang:2016:TRL:3060621.3060801,cdkge}}) learns embedding for both nodes and edges. Each edge is a triplet $<h,r,t>$ (Def. \ref{def:knowgraph}). The embedding is learnt to preserve $r$ between $h$ and $t$ in the embedded space, so that a missing entity/relation can be correctly predicted given the other two components in $<h,r,t>$. 
Secondly, some work (e.g., \cite{node2vec,DBLP:journals/ijmir/ZhaoZWHC16}) embeds a node pair as a vector feature to either make the node pair comparable to other nodes or predict the existence of a link between two nodes. For instance, \cite{DBLP:journals/ijmir/ZhaoZWHC16} proposes a content-social influential feature to predict user-user interaction probability given a content. It embeds both the user pairs and content in the same space. \cite{node2vec} embeds a pair of nodes using a bootstrapping approach over the node embedding, to facilitate the prediction of whether a link exists between two nodes in a graph.

In summary, edge embedding benefits edge (/node pairs) related graph analysis, such as link prediction, knowledge graph entity/relation prediction, etc.

\vspace{0.05in}\noindent
\textbf{Challenge:} \textit{How to define the edge-level similarity and how to model the asymmetric property of the edges, if any?} The edge proximity is different from node proximity as an edge contains a pair of nodes and usually denotes the pairwise node relation. Moreover, unlike nodes, edges may be directed. This asymmetric property should be encoded in the learnt edge representations. 

\subsubsection{Hybrid Embedding}
\label{sec:he}
Hybrid embedding is the embedding of a combination of different types of graph components, e.g, node + edge (i.e., substructure), node + community.

Substructure embedding has been studied in a quantity of work. For example, \cite{DBLP:conf/aaai/LiuZZZCWY17} embeds the graph structure between two possibly distant nodes to support semantic proximity search. \cite{dgk} learns the embedding for subgraphs (e.g., graphlets) so as to define the graph kernels for graph classification. \cite{DBLP:conf/emnlp/BordesCW14} utilizes a knowledge base to enrich the information about the answer. It embeds both path and subgraph from the question entity to the answer entity.

Compared to subgraph embedding, community embedding has only attracted limited attention. \cite{DBLP:conf/aaai/WangCWP0Y17} proposes to consider a community-aware proximity for node embedding, such that a node's embedding is similar to its community's embedding. ComE \cite{DBLP:conf/cikm/CavallariZCCC17} also jointly solves node embedding, community detection and community embedding together. Rather than representing a community as a vector, it defines each community embedding as a multivariate Gaussian distribution so as to characterize how its member nodes are distributed. 

The embedding of substructure or community can also be derived by aggregating the individual node and edge embedding inside it. However, such a kind of ``indirect'' approach is not optimized to represent the structure. Moreover, node embedding and community embedding can reinforce each other. Better node embedding is learnt by incorporating the community-aware high-order proximity, while better communities are detected when more accurate node embedding is generated.

\vspace{0.05in}\noindent
\textbf{Challenge:} \textit{How to generate the target substructure and how to embed different types of graph components in one common space?} In contrast to other types of embedding output, the target to embed in hybrid embedding (e.g., subgraph, community) is not given. Hence the first challenge is how to generate such kind of embedding target structure. Furthermore, different types of targets (e.g., community, node) may be embedded in one common space simultaneously. How to address the heterogeneity of the embedding target types is a problem.

\subsubsection{Whole-Graph Embedding}
\label{sec:wge}
The last type of output is the embedding of a whole graph usually for small graphs, such as proteins, molecules, etc. In this case, a graph is represented as one vector and two similar graphs are embedded to be closer. 

Whole-graph embedding benefits the graph classification task by providing a straightforward and efficient solution for calculating graph similarities \cite{lcnnfg,journals/jmlr/ShervashidzeSLMB11,DBLP:journals/pr/MousaviSMB17}. 
\eat{It is also useful for inferring unseen graph properties such as node types and missing edges \cite{lcnnfg}.} 
To establish a compromise between the embedding time (efficiency) and the ability to preserve information (expressiveness), \cite{DBLP:journals/pr/MousaviSMB17} designs a hierarchical graph embedding framework. \eat{It thinks that although the local processing through checking labels and adjacency relations can provide useful information about a graph,} It thinks that accurate understanding of the global graph information requires the processing of substructures in different scales. A graph pyramid is formed where each level is a summarized graph at different scales. The graph is embedded at all levels and then concatenated into one vector. \cite{DBLP:conf/www/LiMGM17} learns the embedding for a whole cascade graph, and then trains a multi-layer perceptron to predict the increment of the size of the cascade graph in the future.

\vspace{0.05in}\noindent
\textbf{Challenge:} \textit{How to capture the properties of a whole graph and how to make a trade-off between expressiveness and efficiency?} Embedding a whole graph requires capturing the property of a whole graph and is thus more time consuming compared to other types of embedding. The key challenge of whole-graph embedding is how to make a choice between the expressive power of the learnt embedding and the efficiency of the embedding algorithm.

\section{Graph Embedding Techniques}
\label{sec:get}
In this section, we categorize graph embedding methods based on the techniques used. Generally, graph embedding aims to represent a graph in a low dimensional space which preserves as much graph property information as possible. The differences between different graph embedding algorithms lie in how they define the graph property to be preserved. Different algorithms have different insights of the node(/edge/substructure/whole-graph) similarities and how to preserve them in the embedded space. Next, we will introduce the insight of each graph embedding technique, as well as how they quantify the graph property and solve the graph embedding problem.

\subsection{Matrix Factorization}
\label{sec:mf}
Matrix factorization based graph embedding represent graph property (e.g., node pairwise similarity) in the form of a matrix and factorize this matrix to obtain node embedding \cite{gesurvey}. The pioneering studies in graph embedding usually solve graph embedding in this way. In most cases, the input is a graph constructed from non-relational high dimensional data features as introduced in Sec. \ref{sec:cg}. And the output is a set of node embedding (Sec. \ref{sec:ne}). The problem of graph embedding can thus be treated as a structure-preserving dimensionality reduction problem which assumes the input data lie in a low dimensional manifold. There are two types of matrix factorization based graph embedding. One is to factorize \textbf{graph Laplacian eigenmaps}, and the other is to directly factorize \textbf{the node proximity matrix}.

\subsubsection{Graph Laplacian Eigenmaps}
\label{sec:gle}
\textbf{Insight:} \textit{The graph property to be preserved can be interpreted as pairwise node similarities. Thus, a larger penalty is imposed if two nodes with larger similarity are embedded far apart.} 

\begin{table*}[t]
\caption{Graph Laplacian eigenmaps based graph embedding.}
\vspace{-3mm}
\label{tab:mfge}
\centering
\tabcolsep=0.05cm
\scriptsize
\renewcommand{\arraystretch}{1.1}
\begin{tabular} {  l|l|c } \hline
\textbf{GE Algorithm}& $\mathbf{W}$& \textbf{Objective Function} \\ \hline
MDS \cite{DBLP:conf/nips/HofmannB94} & $W_{ij} = $ Euclidean distance $(X_i,X_j)$ & Eq. \ref{eq:mfobj2} \\ \hline
Isomap \cite{Balasubramanian7} & KNN, $W_{ij}$ is the sum of edge weights along the shortest path between $v_i$ and $v_j$ & Eq. \ref{eq:mfobj2}\\ \hline
LE \cite{le85} & KNN, ${W_{ij}} = {\exp({\frac{{{{\| {{X_i} - {X_j}} \|}^2}}}{{2{t^2}}}})}$& Eq. \ref{eq:mfobj2} \\ \hline
LPP \cite{DBLP:conf/nips/HeN03} & KNN, ${W_{ij}} = {\exp({\frac{{{{\| {{X_i} - {X_j}} \|}^2}}}{t}})}$ & Eq. \ref{eq:mfobj4} \\ \hline
AgLPP \cite{DBLP:journals/ijon/JiangFWHH16} &anchor graph, $W=Z\Lambda^{-1} Z^T$, $\Lambda_{kk}=\sum Z_{ik}$, $Z_{ik}=\frac{K_{\sigma}(X_i,U_k)}{\sum_j K_{\sigma}(X_i,U_j)}$&$a^*=\arg\min \frac{{{a^T}UL{U^T}a}}{{{a^T}UD{U^T}a}}$\\\hline
LGRM \cite{DBLP:conf/aaai/YangNXZW10} & KNN, ${W_{ij}} = {\exp({\frac{{{{\| {{X_i} - {X_j}} \|}^2}}}{{2{t^2}}}})}$&$y^* =  \arg \min \frac{{{y^T}(L_{le}+\mu L_g)y}}{{{y^T}y}} $ \\\hline
\multirow{2}{*}{ARE \cite{Lin:2005:SML:1101149.1101193}} &KNN, ${W_{ij}} = \exp({\frac{{{-\rho^2(X_iX_j)}}}{t}})$, $W_{ij}^{ARE} = \left\{ \begin{array}{l}
 - \gamma ,{X_i} \in {F^ + } \& {X_j} \in {F^ + }\\
1,\mathcal{L}(X_i)\ne\mathcal{L}(X_j)\\
0,otherwise
\end{array} \right.$ & \multirow{2}{*}{${a^*} = \arg \max \frac{{{a^T}X{L^{ARE}}{X^T}a}}{{{a^T}XL{X^T}a}}$}\\ 
& $F^+$ denotes the images relevant to a query, $\gamma$ controls the unbalanced feedback\\ \hline
\multirow{2}{*}{SR \cite{DBLP:conf/mm/CaiHH07}} & KNN, ${W_{ij}} = \left\{ \begin{array}{l}
1,\mathcal{L}(X_i)=\mathcal{L}(X_j)\\
0,\mathcal{L}(X_i)\ne\mathcal{L}(X_j)\\
{W_{ij}},otherwise
\end{array} \right.$ $W_{ij}^{SR} = \left\{ \begin{array}{l}
1/{l_r},\mathcal{L}(X_i)=\mathcal{L}(X_j)=C_r\\
0,otherwise
\end{array} \right.$ & \multirow{2}{*}{${a^*} = \arg \max \frac{{{a^T}X{W^{SR}}{X^T}a}}{{{a^T}X({D^{SR}} + L){X^T}a}}$}\\ 
& $C_r$ is the $r$-th class, $l_r$=$|X_i \in X: \mathcal{L}(X_i)=C_r|$\\ \hline
HSL\cite{DBLP:conf/kdd/SunJY08} & $S=I-L$, where $L$ is normalized hypergraph Laplacian & ${a^*} = \arg \max tr({a^T}XS{X^T}a)$, s.t. $a^TXX^Ta=I_k$ \\ \hline
\multirow{2}{*}{MVU \cite{DBLP:conf/icml/WeinbergerSS04}} &KNN, $W^*= \arg\max tr(W)$, s.t. $W \ge 0$,$\sum_{ij} {{W_{ij}} = 0} $ and $\forall i,j$,& \multirow{2}{*}{Eq. \ref{eq:mfobj2}}\\
& where $W_{ii}-2W_{ij}+W_{jj}= \|X_i - X_j \|^2$ & \\\hline
\multirow{2}{*}{SLE \cite{DBLP:conf/aaai/GongTYF14}} &KNN, ${W_{ij}} = \left\{ \begin{array}{l}
1/{l_r},\mathcal{L}(X_i)=\mathcal{L}(X_j)=C_r\\
-1,\mathcal{L}(X_i)\ne\mathcal{L}(X_j)
\end{array} \right.$ & \multirow{2}{*}{Eq. \ref{eq:mfobj4}}\\
&$C_r$ is the $r$-th class, $l_r$=$|X_i \in X: \mathcal{L}(X_i)=C_r|$\\ \hline
\multirow{3}{*}{NSHLRR \cite{DBLP:journals/pami/YinGL16}} & normal graph: KNN, ${W_{ij}} = 1$& Eq. \ref{eq:mfobj2}\\\cline{2-3}
 &hypergraph: $W(\mathbf{e})$ is the weight of a hyperedge $\mathbf{e}$& \multirow{2}{*}{${y^*} = \arg \min \sum\limits_{\mathbf{e}} {{{\| {{y_i} - {y_j}} \|}^2}\frac{{W(\mathbf{e})}}{{d(\mathbf{e})}}} $}\\
& $h(v,\mathbf{e}) = \left\{ \begin{array}{l}
1,v \in \mathbf{e}\\
0,otherwise
\end{array} \right.$,  $d(\mathbf{e}) = \sum\nolimits_{v \in \mathcal{V}} {h(v,e)}$&\\\hline
\cite{DBLP:conf/mm/TangNJ16}&${W_{ij}} = \left\{ \begin{array}{l}
\frac{{\| {{X_i} - {X_{m + 1}}} \|_2^2 - \| {{X_i} - {X_j}} \|_2^2}}{{k\| {{X_i} - {X_{k + 1}}} \|_2^2 - \sum_{m = 1}^k {\| {{X_i} - {X_k}} \|_2^2} }},j \le k\\
0,j > k
\end{array} \right.$&${y^*} = \arg \mathop {\min }\limits_{{y^T}y = 1} \sum\limits_{i \ne j} {{W_{ij}}\min (\| {{y_i} - {y_j}} \|_2^p,\theta )} $\\\hline
PUFS \cite{DBLP:conf/ijcai/HanS16} &KNN, ${W_{ij}} = {\exp({-\frac{{{{\| {{X_i} - {X_j}} \|}^2}}}{{2{t^2}}}})}$&Eq. \ref{eq:mfobj4} +(must-link and cannot link constraints) \\\hline
RF-Semi-NMF-PCA \cite{DBLP:journals/tkde/AllabLN17}&KNN , ${W_{ij}} = 1$ & Eq. \ref{eq:mfobj2} +$\mathcal{O}$(PCA) + $\mathcal{O}$(kmeans) \\\hline
\end{tabular}
\vspace{-1mm}
\end{table*}

Based on the above insight, the optimal embedding $y$ can be derived by the below objective function \cite{DBLP:conf/mm/CaiHH07}.
\vspace{-1mm}
\begin{equation} \label{eq:mfobj1}
{y^*} = \arg \min \sum\nolimits_{i \ne j} ({{y_i} - {y_j}{^2}{W_{ij}}})  = \arg \min {y^T}Ly,
\end{equation}
where $W_{ij}$ is the ``defined'' similarity between node $v_i$ and $v_j$; $L\!=\!D\!-\!W$ is the graph Laplacian. $D$ is the diagonal matrix where ${D_{ii}} = \sum\nolimits_{j \ne i} {{W_{ij}}}$. The bigger the value of $D_{ii}$, the more important $y_i$ is \cite{DBLP:conf/nips/HeN03}. A constraint $y^TDy=1$ is usually imposed on Eq. \ref{eq:mfobj1} to remove an arbitrary scaling factor in the embedding.
Eq. \ref{eq:mfobj1} then reduces to:
\vspace{-2mm}
\begin{equation} \label{eq:mfobj2}
 y^* = \arg \min \limits_{{y^T}Dy = 1} {y^T}Ly = \arg \min \frac{{{y^T}Ly}}{{{y^T}Dy}} =  \arg \max \frac{{{y^T}Wy}}{{{y^T}Dy}}.
 \end{equation}
 The optimal $y$'s are the eigenvectors corresponding to the maximum eigenvalue of the eigenproblem $Wy = \lambda Dy$.

The above graph embedding is \textit{transductive} because it can only embed the nodes that exist in the training set. In practice, it might also need to embed the new coming nodes that have not been seen in training. One solution is to design a linear function $y=X^Ta$ so that the embedding can be derived as long as the node feature is provided. Consequently, for \textit{inductive} graph embedding, Eq. \ref{eq:mfobj1} becomes finding the optimal $a$ in the below objective function:
\vspace{-1mm}
\begin{equation} \label{eq:mfobj3}
a^* \!=\! \arg \min \sum\nolimits_{i \ne j} {{{\| \!{{a^T}\!{X_i} \!-\! {a^T}\!{X_j}}\!\|}^2}\!{W_{ij}} \!=\!\arg \min {a^T}\!XL{X^T}\!a}\!.
\end{equation}

Similar to Eq. \ref{eq:mfobj2}, by adding the constraint $a^TXDX^Ta=1$, the problem in Eq. \ref{eq:mfobj3} becomes:
\begin{equation} \label{eq:mfobj4}
%\vspace{-2mm}
a^* \!=\! \arg\min \frac{{{a^T}XL{X^T}a}}{{{a^T}XD{X^T}a}} = \arg \max \frac{{{a^T}XW{X^T}a}}{{{a^T}XD{X^T}a}}.
\end{equation}
The optimal $a$'s are eigenvectors with the maximum eigenvalues in solving $XW{X^T}a = \lambda XD{X^T}a$.

The differences of existing studies mainly lie in how they calculate the pairwise node similarity $W_{ij}$, and whether they use a linear function $y=X^Ta$ or not. Some attempts \cite{DBLP:journals/pami/YanXZZYL07,DBLP:journals/tkde/ChenTTJ15} have been made to summarize existing Laplacian eigenmaps based graph embedding methods using a general framework. But their surveys only cover a limited quantity of work. 
In Table \ref{tab:mfge}, we summarize existing Laplacian eigenmaps based graph embedding studies and compare how they calculate $W$ and what objective function they adopt.

The initial study MDS\cite{DBLP:conf/nips/HofmannB94} directly adopted the Euclidean distance between two feature vectors $X_i$ and $X_j$ as $W_{ij}$. Eq. \ref{eq:mfobj2} is used to find the optimal embedding $y$'s. MDS does not consider the neighbourhood of nodes, i.e., any pair of training instances are considered as connected. The follow-up studies (e.g., \cite{Balasubramanian7,Roweis2000,le85,DBLP:conf/nips/HeN03}) overcome this problem by first constructing a k nearest neighbour (KNN) graph from the data feature. Each node is only connected with its top k similar neighbours. After that, different methods are utilized to calculate the similarity matrix $W$ so as to preserve as much desired graph property as possible. Some more advanced models are design recently. For example, AgLPP \cite{DBLP:journals/ijon/JiangFWHH16} introduces an anchor graph to significantly improve the efficiency of earlier matrix factorization model LPP. LGRM \cite{DBLP:conf/aaai/YangNXZW10} learns a local regression model to grasp the graph structure and a global regression term for out-of-sample data extrapolation. Finally, different from previous work’s preserving local geometry, LSE \cite{DBLP:journals/tkde/XiangNZZ09} uses local spline regression to preserve global geometry. 

When auxiliary information (e.g., label, attribute) is available, the objective function is adjusted to preserve the richer information. E.g., \cite{DBLP:conf/mm/CaiHH07} constructs an adjacency graph $W$ and a labelled graph $W^{SR}$. The objective function consists of two parts, one focuses on preserving the local geometric structure of the datasets as in LPP \cite{DBLP:conf/nips/HeN03}, and the other tries to get the embedding with the best class separability on the labelled training data. Similarly, \cite{Lin:2005:SML:1101149.1101193} also constructs two graphs, an adjacency graph $W$ which encodes local geometric structures and a feedback relational graph $W^{ARE}$ that encodes the pairwise relations in users' relevance feedbacks. RF-Semi-NMF-PCA \cite{DBLP:journals/tkde/AllabLN17} simultaneously consider clustering, dimensionality reduction and graph embedding by constructing an objective function that consists of three components: PCA, k-means and graph Laplacian regularization. 

Some other work thinks that $W$ cannot be constructed by easily enumerating pairwise node relationships. Instead, they adopt semidefinite programming (SDP) to learn $W$. Specifically, SDP \cite{Vanden1996} aims to find an inner product matrix that maximizes the pairwise distances between any two inputs which are not connected in the graph while preserving the nearest neighbors distances. MVU \cite{DBLP:conf/icml/WeinbergerSS04} constructs such matrix and then applies MDS \cite{DBLP:conf/nips/HofmannB94} on the learned inner product matrix. 
\cite{DBLP:conf/aaai/NieZL17} proves that regularized LPP \cite{DBLP:conf/nips/HeN03} is equivalent to regularized SR \cite{DBLP:conf/mm/CaiHH07} if $W$ is symmetric, doubly stochastic, PSD and with rank $p$. It constructs such kind of similarity matrix so as to solve LPP liked problem efficiently.

\begin{table*}[t]
\caption{Node proximity matrix factorization based graph embedding. $\mathcal{O}$($*$) denotes the objective function; \emph{e.g.}, $\mathcal
O$(SVM classifier) denotes the objective function of a SVM classifier.}.
\vspace{-5mm}
\label{tab:mfge2}
\centering
\tabcolsep=0.05cm
\scriptsize
\renewcommand{\arraystretch}{1.1}
\begin{tabular} {  l|l|c } \hline
\textbf{GE Algorithm}& $\mathbf{W}$& \textbf{Objective Function} \\ \hline
\cite{DBLP:conf/aaai/NikolentzosMV17}&${W_{ij}} = \left\{ \begin{array}{l}
1,{e_{ij}} \in E\\
0,otherwise
\end{array} \right.$& Eq. \ref{eq:mf}\\ \hline
\multirow{2}{*}{SPE\cite{DBLP:conf/icml/ShawJ09}} &KNN, ${W^*} = \arg \mathop {\max }\limits_{k \ge 0} tr(W\hat{A})$, s.t. ${D_{ij}} > (1 - {\hat{A}_{ij}}){\max\limits _m}({\hat{A}_{im}}{D_{im}})$& \multirow{2}{*}{Eq. \ref{eq:mf}} \\
& $\hat{A}$ is a connectivity matrix that describes local pairwise distances & \\\hline
\multirow{2}{*}{HOPE \cite{atpge}} & Katz Index $W=(\mathbf{I}-\beta A)^{-1}\beta A$; Personalized Pagerank $W=(1-\alpha)(\mathbf{I}-\alpha P)^{-1}$&\multirow{2}{*}{Eq. \ref{eq:mf}} \\ 
& Common neighbors $W=A^2$; Adamic-Adar $W=A (1/ \sum_j (A_{ij}+A_{ji}))A$& \\\hline
GraRep \cite{grarep} & $W_{ij}^k = \log (\frac{\hat{A}_{ij}^k}{\sum_t \hat{A}_{ij}^k})-\log (\frac{\lambda}{|V|})$, where $\hat{A}=D^{-1}S$, ${D_{ij}} = \left\{ \begin{array}{l}
\sum_p {{A_{ip}}} ,i = j\\
0,i \ne j
\end{array} \right.$& Eq. \ref{eq:mf}\\\hline
CMF \cite{rlfmerri} & PPMI & Eq. \ref{eq:mf}\\\hline
TADW \cite{nrlrti} & PMI& Eq. \ref{eq:mf} with text feature matrix\\\hline
\cite{Ahmed:2013:DLN:2488388.2488393}& A& ${y^*} = \arg \min \sum_{{e_{ij}} \in E} {{{({A_{ij}} -  < {y_i},{y_j} > )}^2}}  + \frac{\lambda }{2}\sum_i {{{\| {{y_i}} \|}^2}} $ \\\hline
MMDW \cite{DBLP:conf/ijcai/TuZLS16} & PMI &Eq. \ref{eq:mf} + $\mathcal{O}$(SVM classifier) \\ \hline
HSCA \cite{DBLP:conf/icdm/ZhangYZZ16} & PMI& $\mathcal{O}$(MMDW) + ($1^{st}$ order proximity constraint) \\\hline
\multirow{2}{*}{MVE \cite{DBLP:journals/jmlr/ShawJ07}} &KNN, ${W^*} = \arg \min tr(W(\sum_{i = 1}^d {{\upsilon _i}\upsilon _i^T}  + \sum_{i = d + 1}^N {{\upsilon _i}\upsilon _i^T} ))$& \multirow{2}{*}{Eq. \ref{eq:mf}}\\
&where $\upsilon _i$ is eigenvector of a pairwise distance matrix, $d$ is embedding dimensionality&\\\hline
\multirow{2}{*}{M-NMF \cite{DBLP:conf/aaai/WangCWP0Y17}} & $W = s^{(1)} + 5s^{(2)}$  &  Eq. \ref{eq:mf} + $\mathcal{O}$(community detection) \\
& &+ (community proximity constraint)\\\hline
\multirow{2}{*}{ULGE \cite{DBLP:conf/aaai/NieZL17}}&$W = Z{\Delta ^{ - 1}}{Z^T}$, where ${Z^*} =  \mathop {\arg\min }\limits_{z_i^T1 = 1,{z_i} \ge 0} \sum_{j = 1}^m {\| {{X_i} - {u_j}} \|_2^2{z_{ij}}}  + \gamma \sum_{j = 1}^m {z_{ij}^2} $ & ${a^*} = \arg \min \| {{a^T}X - {F_p}} \|_F^2 + \alpha \| a \|_F^2$ \\
&& where$F_p$ is the top $p$ eigenvectors of $W$ \\\hline
LLE \cite{Roweis2000} & KNN, $W^* = \arg\min \sum_i {\| X_i - \sum_j{W_{ij}X_j} \|^2}$ & $y^* =\arg\min \sum_i{ \| y_i - \sum_j{W_{ij}y_j} \|^2}$\\ \hline 
RESCAL \cite{ICML2011Nickel_438} & ${W_{ijk}} = \left\{ \begin{array}{l}
1,({h_i},{r_j},{t_k})\text{ exists}\\
0,otherwise
\end{array} \right.$& $\min \sum\limits_k {\| {{W_k} - Y{R_k}{Y^T}} \|_F^2}  + \lambda (\| Y \|_F^2 + \sum\limits_k {\| {{R_k}} \|} _F^2)$ \\\hline
FONPE \cite{ijcai2017-361} & KNN, $W^* = \arg\min \sum_i {\| X_i - \sum_j{W_{ij}X_j} \|^2}$ & $\min \| {F - F{W^T}} \|_F^2 + \beta \| {{P^T}X - F} \|_F^2$, s.t. $P^TP=\mathbf{I}$\\ \hline 
\end{tabular}
%\vspace{-1mm}
\end{table*} 

\subsubsection{Node Proximity Matrix Factorization}
\label{sec:npmf}
In addition to solving the above generalized eigenvalue problem, another line of studies tries to directly factorize node proximity matrix.

\textbf{Insight:} \textit{Node proximity can be approximated in a low-dimensional space using matrix factorization. The objective of preserving node proximity is to minimize the loss of approximation.}

Given the node proximity matrix $W$, the objective is:
\vspace{-1mm}
\begin{equation}
\label{eq:mf}
\textstyle
\min  \| W- Y{Y^c}^T  \|,
\vspace{-1mm}
\end{equation}
where $Y \in \mathbb{R}^{|V|\times d}$ is the node embedding, and $Y^c \in \mathbb{R}^{|V|\times d}$ is the embedding for the context nodes \cite{grarep}.

Eq. \ref{eq:mf} aims to find an optimal rank-$d$ approximation of the proximity matrix $W$ ($d$ is the dimensionality of the embedding). One popular solution is to apply SVD (Singular Value Decomposition) on $W$ \cite{Golub:1970:SVD:2717826.2718152}. Formally,
\vspace{-1mm}
\begin{equation}
\textstyle
W = \sum_{i=1}^{|V|} \sigma_iu_i{u_i^c}^T  \approx \sum_{i=1}^{d} \sigma_iu_i{u_i^c}^T, 
\end{equation}
where $\{\sigma_1, \sigma_2, \cdots, \sigma_{|V|}\}$ are the singular values sorted in descending order, $u_i$ and $u_i^c$ are singular vectors of $\sigma_i$. The optimal embedding is obtained using the largest $d$ singular values and corresponding singular vectors as follows: 
\vspace{-1mm}
\begin{equation}
\textstyle
\begin{aligned}
Y=[\sqrt{\sigma_1}u_1, \cdots, \sqrt{\sigma_d}u_d], \\
Y^c=[\sqrt{\sigma_1}u_1^c, \cdots, \sqrt{\sigma_d}u_d^c].
\end{aligned}
\end{equation}

Depending on whether the asymmetric property is preserved or not, the embedding of node $i$ is either $y_i = Y_i$ \cite{grarep,DBLP:conf/aaai/NikolentzosMV17}, or the concatenation of $Y_i$ and $Y^c_i$, i.e., $y_i = [ Y_i, Y^c_i]$ \cite{atpge}. There exist other solutions for Eq. \ref{eq:mf}, such as regularized Gaussian matrix factorization \cite{Ahmed:2013:DLN:2488388.2488393}, low-rank matrix factorization \cite{nrlrti}, and adding other regularizers to enforce more constraints \cite{DBLP:conf/ijcai/TuZLS16}.
We summarize all the node proximity matrix factorization based graph embedding in Table \ref{tab:mfge2}.

\vspace{0.05in}\noindent
\textbf{Summary:} Matrix Factorization (MF) is mostly used to embed a graph constructed from non-relational data (Sec. \ref{sec:cg}) for node embedding (Sec. \ref{sec:ne}), which is the typical setting of graph Laplacian eigenmap problems. MF is also used to embed homogeneous graphs \cite{DBLP:conf/aaai/NikolentzosMV17, Ahmed:2013:DLN:2488388.2488393} (Sec. \ref{sec:hog}).  

\subsection{Deep Learning}
\label{sec:dl}
Deep learning (DL) has shown outstanding performance in a wide variety of research fields, such as computer vision, language modeling, etc.
DL based graph embedding applies DL models on graphs.
These models are either a direct adoption from other fields or a new neural network model specifically designed for embedding graph data. 
The input is either paths sampled from a graph or the whole graph itself. Consequently, we divide the DL based graph embedding into two categories based on whether random walk is adopted to sample paths from a graph.

\subsubsection{DL based Graph Embedding with Random Walk}
\textbf{Insight:} \textit{The second-order proximity in a graph can be preserved in the embedded space by maximizing the probability of observing the neighbourhood of a node conditioned on its embedding.}

In the first category of deep learning based graph embedding, a graph is represented as a set of random walk paths sampled from it. The deep learning methods are then applied to the sampled paths for graph embedding which preserves graph properties carried by the paths.

In view of the above insight, DeepWalk \cite{deepwalk} adopts a neural language model (SkipGram) for graph embedding. SikpGram \cite{skipgram} aims to maximize the co-occurrence probability among the words that appear within a window $w$. DeepWalk first samples a set of paths from the input graph using truncated random walk (i.e., uniformly sample a neighbour of the last visited node until the maximum length is reached). Each path sampled from the graph corresponds to a sentence from the corpus, where a node corresponds to a word. Then SkipGram is applied on the paths to maximize the probability of observing a node's neighbourhood conditioned on its embedding. In this way, nodes with similar neighbourhoods (having large second-order proximity values) share similar embedding. The objective function of DeepWalk is as follows:
\vspace{-1mm}
\begin{equation}
\label{eq:skipgram}
\textstyle
\mathop {\min }_y  - \log P(\{ {v_{i - w}}, \cdots ,{v_{i - 1}},{v_{i + 1}}, \cdots ,{v_{i + w}}\} |{y_i}),
\end{equation}
where $w$ is the window size which restricts the size of random walk context. SkipGram removes the ordering constraint, and Eq. \ref{eq:skipgram} is transformed to:
\vspace{-1mm}
\begin{equation}
\label{eq:skipgram2}
\textstyle
\mathop {\min }_y  - \log \sum_{ - w \le j \le w} {P({v_{i + j}}|{y_i})},
\end{equation}
where $ P({v_{i + j}}|{y_i})$ is defined using the softmax function:
\vspace{-1mm}
\begin{equation}
\label{eq:softmax}
\textstyle
P({v_{i + j}}|{y_i}) = \frac{{\exp (y_{i + j}^T{y_i})}}{{\sum_{k = 1}^{|V|} {\exp (y_k^T{y_i})} }}.
\end{equation}
Note that calculating Eq. \ref{eq:softmax} is not feasible as the normalization factor (i.e., the summation over all inner product with every node in a graph) is expensive. There are usually two solutions to approximate the full softmax: hierarchical softmax \cite{DBLP:conf/nips/MikolovSCCD13} and negative sampling \cite{DBLP:conf/nips/MikolovSCCD13}. 

\textbf{Hierarchical softmax}: To efficiently solve Eq. \ref{eq:softmax}, a binary tree is constructed in which the nodes are assigned to the leaves. Instead of enumerating all nodes as in Eq. \ref{eq:softmax}, only the path from the root to the corresponding leaf needs to be evaluated. The optimization problem becomes maximizing the probability of a specific path in the tree. Suppose the path to leaf $v_i$ is a sequence of nodes $(b_0, b_1, \cdots, b_{log(|V|)})$, where $b_0 =$ root, $b_{log(|V|)} = v_i$. Eq. \ref{eq:softmax} then becomes:
\vspace{-1mm} 
\begin{equation}
\label{eq:hs}
\textstyle
P({v_{i + j}}|{y_i}) = \prod_{t = 1}^{\log (|V|)} {P({b_t}|{y_i})},
\end{equation}
where $P({b_t})$ is a binary classifier:
$P({b_t}|{v_i}) = \sigma (y_{{b_t}}^T{y_i})$. $\sigma(\cdot)$ denotes the sigmoid function. $y_{{b_t}}$ is the embedding of tree node $b_t$'s parent. The hierarchical softmax reduces time complexity of SkipGram from $\mathcal{O}(|V|^2)$ to $\mathcal{O}(|V|log(|V|))$.

\textbf{Negative sampling}: The key idea of negative sampling is to distinguish the target node from noises using logistic regression. I.e., for a node $v_i$, we want to distinguish its neighbour $v_{i+j}$ from other nodes. A noise distribution  $P_{n}(v_i)$ is designed to draw the negative samples for node $v_i$. 
Each $\log P({v_{i + j}}|{y_i})$ in Eq. \ref{eq:skipgram2} is then calculated as:
\vspace{-1mm}
\begin{equation}
\textstyle
\label{eq:ns}
\log \sigma (y_{i + j}^T{y_i}) + \sum_{t = 1}^K {{E_{{v_t} \sim {P_n}}}[\log \sigma ( - y_{{v_t}}^T{y_i})]}, 
\end{equation}
where $K$ is the number of negative nodes that are sampled. $P_n(v_i)$ is a noise distribution, e.g., a uniform distribution ($P_n(v_i) \sim \frac{1}{|V|}, \forall v_i \in V$). The time complexity of SkipGram with negative sampling is $\mathcal{O}(K|V|)$.

\eat{In contrast to hierarchical softmax which reduces the time complexity by reducing the optimization space from a whole tree to a path in the tree, negative sampling update Eq. \ref{eq:skipgram2} using a set of negative samples rather than the whole node set. , where $K$ is the number of sampled negative nodes.}

%%%%%
\begin{table}[t]
\caption{Deep Learning based graph embedding \emph{with} random walk paths.}
\vspace{-3mm}
\label{tab:dlgerw}
\centering
\tabcolsep=0.05cm
\scriptsize
\renewcommand{\arraystretch}{1.2}
\begin{tabular} {  p{1.9cm}|p{3cm}|p{2.3cm}|p{1.4cm} } \hline
\textbf{GE Algorithm}& \textbf{Ransom Walk Methods}& \textbf{Preserved Proximity} & \textbf{DL Model} \\ \hline
DeepWalk\cite{deepwalk}&  truncated random walk & $2^{nd}$ \\ \cline{1-3}
\cite{DBLP:journals/tomccap/ZhangSLWC16} &  truncated random walk & $2^{nd}$ (word-image)&  \\ \cline{1-3}
GenVector \cite{mmbeflskg}&  truncated random walk & $2^{nd}$ (user-user \& concept-concept)& SkipGram with  \\ \cline{1-3}
Constrained DeepWalk \cite{DBLP:conf/ijcnn/JinLLZZW16}& sampling with edge weight & $2^{nd}$ & hierarchical softmax \\ \cline{1-3}
DDRW \cite{DBLP:conf/acl/LiZZ16} & truncated random walk & $2^{nd}$ + class identity  & (Eq. \ref{eq:hs})\\ \cline{1-3}
TriDNR \cite{DBLP:conf/ijcai/PanWZZW16} & truncated random walk & $2^{nd}$ (among node, word \& label)  \\ \hline
node2vec \cite{node2vec} & BFS + DFS & $2^{nd}$ &  \\ \cline{1-3}
UPP-SNE \cite{ijcai2017-472}&  truncated random walk & $2^{nd}$ (user-user \& profile-profile) & SkipGram with \\ \cline{1-3}
Planetoid \cite{rsslge} & sampling node pairs by labels and structure & $2^{nd}$ + label identity & negative sampling \\ \cline{1-3}
NBNE \cite{tpavnz} & sampling direct neighbours of a node & $2^{nd}$ & (Eq. \ref{eq:ns}) \\\hline
DGK \cite{dgk} &graphlet kernel: random sampling \cite{DBLP:journals/jmlr/ShervashidzeVPMB09} & $2^{nd}$ (by graphlet) &SkipGram (Eqs. \ref{eq:hs}--\ref{eq:ns} )\\ \hline
metapath2vec \cite{dong2017metapath2vec} & meta-path based random walk & $2^{nd}$ & heterogeneous SkipGram \\ \hline
ProxEmbed \cite{DBLP:conf/aaai/LiuZZZCWY17} & truncate random walk &  node ranking tuples &\\ \cline{1-3}
HSNL \cite{cqahsnl} & truncate random walk & $2^{nd}$ + QA ranking tuples &LSTM \\ \cline{1-3}
RMNL \cite{DBLP:conf/ijcai/ZhaoYCHZ16} & truncated random walk& $2^{nd}$ + user-question quality ranking  &\\ \hline
DeepCas \cite{DBLP:conf/www/LiMGM17}&Markov chain based random walk &information cascade sequence &GRU \\ \hline
MRW-MN \cite{DBLP:journals/tip/WuLSYZRZ16}&truncated random walk&$2^{nd}$ + cross-modal feature difference&DCNN+ SkipGram\\ \hline
\end{tabular}
\vspace{-1mm}
\end{table}
%%%%%

The success of DeepWalk \cite{deepwalk} motivates many subsequent studies which apply deep learning models (e.g., SkipGram or Long-Short Term Memory (LSTM) \cite{lstm}) on the sampled paths for graph embedding. We summarize them in Table \ref{tab:dlgerw}\eat{, and compare them in terms of their choices of random walk methods, the proximity they preserve and the models they use}. 
As shown in the table, most studies follow the idea of DeepWalk but change the settings of either random walk sampling methods (\cite{DBLP:conf/ijcnn/JinLLZZW16,node2vec,rsslge,rsslge}) or proximity (Def. \ref{def:fpro} and Def. \ref{def:spro}) to be preserved (\cite{DBLP:journals/tomccap/ZhangSLWC16,mmbeflskg,DBLP:conf/acl/LiZZ16,DBLP:conf/ijcai/PanWZZW16,rsslge}). \eat{DGK \cite{dgk} embeds different structures to provide deep graph kernels.}  \cite{dong2017metapath2vec} designs meta-path-based random walks to deal with heterogeneous graphs and  a heterogeneous SkipGram \eat{($\arg \mathop {\max }\limits_\theta  \sum\nolimits_{v \in V} {\sum\nolimits_{t \in {\mathcal{T}^v}} {\sum\nolimits_{{c_t} \in {N_t}(v)} {\log p({c_t}|v;\theta )} } } $)} which maximizes the probability of having the hetegeneous context for a given node. Apart from SkipGram, LSTM is another popular deep learning model adopted in graph embedding. Note that SkipGram can only embed one single node. However, sometimes we may need to embed a sequence of nodes as a fixed length vector, e.g., represent a sentence (i.e., a sequence of words) as one vector. LSTM is then adopted in such scenarios to embed a node sequence. For example, \cite{cqahsnl} and \cite{DBLP:conf/ijcai/ZhaoYCHZ16} embed the sentences from questions/answers in cQA sites, and \cite{DBLP:conf/aaai/LiuZZZCWY17} embeds a sequence of nodes between two nodes for proximity embedding. A ranking loss function is optimized in these work to preserve the ranking scores in the training data. In \cite{DBLP:conf/www/LiMGM17}, GRU \cite{gru} (i.e., a recurrent neural network model similar to LSTM) is used to embed information cascade paths.
\eat{Given a sequence of nodes $\{v_1, v_2, \cdots, v_k\}$ as the input, the recurrent neural network model LSTM learns the embedding for the sequence by the following equations:

\begin{equation}
\label{eq:lstm}
\begin{aligned}
&i_t = \sigma (W_iv_i + U_ih_{t-1}+b_i), \\
&\hat{C_t} = \tanh(W_cv_t+U_fh_{t-1}+b_f),\\
&f_t=\sigma(W_fv_t+U_fh_{t-1}+b_f),\\
&C_t=i_t\hat{C_t}+f_tC_{t-1},\\
&o_t=\sigma(W_0v_t+U_0h_{t-1}+V_0C_t+b_0),\\
&h_t=o_t\tanh(C_t).
\end{aligned}
\end{equation}
where $\sigma$ is the sigmoid activation function, $W$, $U$ and $V$ are weight matrices, $b$ is the bias vector. The output of the last LSTM cell ($h_k$) is taken as the embedding of the sequence $\{v_1, v_2, \cdots, v_k\}$.

Based on Eq. \ref{eq:lstm},} 
\subsubsection{DL based Graph Embedding without Random Walk}
\textbf{Insight:}\textit{The multi-layered learning architecture is a robust and effective solution to encode the graph into a low dimensional space.}

The second class of deep learning based graph embedding methods applies deep models on a whole graph (or a proximity matrix of a whole graph) directly. Below are some popular deep learning models used in graph embedding.

\textbf{Autoencoder}: An autoencoder aims to minimize the reconstruction error of the output and input by its encoder and decoder. Both encoder and decoder contain multiple nonlinear functions. The encoder maps input data to a representation space and the decoder maps the representation space to a reconstruction space. The idea of adopting autoencoder for graph embedding is similar to node proximity matrix factorization (Sec. \ref{sec:npmf}) in terms of neighbourhood preservation. Specifically, the adjacency matrix captures a node's neighbourhood. If we input the adjacency matrix to an autoencoder, the reconstruction process will make the nodes with similar neighbourhood have similar embedding.

\textbf{Deep Neural Network}: As a popular deep learning model, Convolutional Neural Network (CNN) and its variants have been widely adopted in graph embedding. On the one hand, some of them directly use the original CNN model designed for Euclidean domains and reformat input graphs to fit it. E.g., \cite{lcnnfg} uses graph labelling to select a fixed-length node sequence from a graph and then assembles nodes' neighbourhood to learn a neighbourhood representation with the CNN model. On the other hand, some other work attempts to generalize the deep neural model to non-Euclidean domains (e.g., graphs). \cite{DBLP:journals/corr/BronsteinBLSV16} summarizes the representative studies in their survey. Generally, the differences between these approaches lie in the way they formulate a convolution-like operation on graphs. One way is to emulate the Convolution Theorem to define the convolution in the spectral domain \cite{DBLP:journals/corr/BrunaZSL13, DBLP:journals/corr/HenaffBL15}. Another is to treat the convolution as neighborhood matching in the spatial domain \cite{DBLP:conf/nips/DefferrardBV16,DBLP:journals/corr/KipfW16,DBLP:journals/tnn/ScarselliGTHM09}. 

\textbf{Others}:
There are some other types of deep learning based graph embedding methods. E.g., 
\cite{DBLP:conf/iccv/GengZBC15} proposes DUIF, which uses a hierarchical softmax as a forward propagation to maximize the modularity. HNE \cite{hneda} utilizes deep learning techniques to capture the interactions between heterogeneous components, e.g., CNN for image and FC layers for text. ProjE \cite{DBLP:conf/aaai/ShiW17} designs a neural network with a combination layer and a projection layer. It defines a pointwise loss (similar to multi-class classification) and a listwise loss (i.e., softmax regression loss) for knowledge graph embedding.

We summarize all deep learning based graph embedding methods (random walk free) in Table \ref{tab:dlgerw2}, and compare the models they use as well as the input for each model.

\begin{table}[t]
\caption{Deep learning based graph embedding \emph{without} random walk paths.}
\vspace{-3mm}
\label{tab:dlgerw2}
\centering
\tabcolsep=0.05cm
\scriptsize
\renewcommand{\arraystretch}{1.1}
\begin{tabular} {  l|l|l } \hline
\textbf{GE Algorithm}&  \textbf{Deep Learning Model}  & \textbf{Model Input}\\ \hline
SDNE \cite{sdne} & autoencoder & $A$ \\\hline
DNGR \cite{dnnflgr} & stacked denoising autoencoder &PPMI\\\hline
SAE \cite{DBLP:conf/aaai/TianGCCL14} &sparse autoencoder& $D^{-1}A$ \\\hline
\cite{lcnnfg} & CNN & node sequence \\\hline
SCNN \cite{DBLP:journals/corr/BrunaZSL13} & Spectral CNN & graph \\\hline
\cite{DBLP:journals/corr/HenaffBL15} & Spectral CNN with smooth&graph \\
& spectral multipliers & \\\hline
MoNet \cite{DBLP:journals/corr/MontiBMRSB16} & Mixture model network & graph \\\hline
ChebNet \cite{DBLP:conf/nips/DefferrardBV16} & Graph CNN a.k.a. ChebNet & graph \\\hline
GCN \cite{DBLP:journals/corr/KipfW16} & Graph Convolutional Network & graph \\\hline
GNN \cite{DBLP:journals/tnn/ScarselliGTHM09} & Graph Neural Network & graph \\\hline
\cite{Duvenaud:2015:CNG:2969442.2969488} & adapted Graph Neural Network & molecules graph  \\\hline
GGS-NNs \cite{DBLP:journals/corr/LiTBZ15} & adapted Graph Neural Network & graph \\\hline
HNE \cite{hneda} & CNN + FC & graph with image and text \\ \hline
DUIF \cite{DBLP:conf/iccv/GengZBC15} & a hierarchical deep model & social curation network \\ \hline
ProjE \cite{DBLP:conf/aaai/ShiW17} & a neural network model & knowledge graph \\\hline
TIGraNet \cite{pmlr-v70-khasanova17a}& Graph Convolutional Network & graph constructed from images\\\hline
\end{tabular}
\vspace{-1mm}
\end{table}

\vspace{0.05in}\noindent
\textbf{Summary:} Due to its robustness and effectiveness, deep learning has been widely used in graph embedding. Three types of input graphs (except for graph constructed from non-relational data (Sec. \ref{sec:cg})) and all the four types of embedding output have been observed in deep learning based graph embedding methods.

\subsection{Edge Reconstruction based Optimization}
\label{sec:ml}
\textbf{Overall Insight:} \textit{The edges established based on node embedding should be as similar to those in the input graph as possible.}

The third category of graph embedding techniques directly optimizes an edge reconstruction based objective functions, by either maximizing edge reconstruction probability or minimizing edge reconstruction loss. The later is further divided into distance-based loss and margin-based ranking loss. Next, we introduce the three types one by one.

\subsubsection{Maximizing Edge Reconstruction Probability}
\label{sec:merp}
\textbf{Insight:} \textit{Good node embedding maximizes the probability of generating the observed edges in a graph.}

Good node embedding should be able to re-establish edges in the original input graph. This can be realized by maximizing the probability of generating all observed edges (i.e., node pairwise proximity) using node embedding.

The direct edge between a node pair $v_i$ and $v_j$ indicates their \textit{first-order proximity}, which can be calculated as the joint probability using the embedding of $v_i$ and $v_j$:
\vspace{-1mm}
\begin{equation}
\label{eq:pfp}
\textstyle
{p^{(1)}}({v_i},{v_j}) = \frac{1}{{1 + \exp ( - {y_i}^T{y_j})}}.
\end{equation}
The above first-order proximity exists between any pair of connected nodes in a graph. To learn the embedding, we maximize the log-likelihood of observing these proximities in a graph. The objective function is then defined as:
\vspace{-1mm}
\begin{equation}
\textstyle
\label{eq:mlfo1}
\mathcal{O}_{max}^{(1)} = \mathop {\max } \sum_{{e_{ij}} \in E} {\log {p^{(1)}}({v_i},{v_j})}. 
\end{equation}
Similarly, \textit{second-order proximity} of $v_i$ and $v_j$ is the conditional probability of $v_j$ generated by $v_i$ using $y_i$ and $y_j$:
\vspace{-1mm}
\begin{equation}
\textstyle
\label{eq:psp}
{p^{(2)}}({v_j}|{v_i}) = \frac{\exp(y_j^T{y_i})}{{\sum_{k = 1}^{|V|} {\exp (y_k^T{y_i})} }}.
\end{equation}
It can be interpreted as the probability of a random walk in a graph which starts from $v_i$ and ends with $v_j$. Hence the graph embedding objective function is: 
\vspace{-1mm}
\begin{equation}
\label{eq:mlso1}
\textstyle
\mathcal{O}_{max}^{(2)} = \mathop {\max }  \sum_{{\{v_i, v_j\}} \in \mathcal{P}} {\log {p^{(2)}}({v_j}|{v_i})}, 
\end{equation}
where $\mathcal{P}$ is a set of $\{ start\_node, end\_node \}$ in the paths sampled from the graph, i.e., the two end nodes from each sampled path. This simulates the second-order proximity as the probability of a random walk starting from the $start\_node$ and ending with the $end\_node$.

\subsubsection{Minimizing Distance-based Loss}
\label{sec:mdbl}
\textbf{Insight:} \textit{The node proximity calculated based on node embedding should be as close to the node proximity calculated based on the observed edges as possible.}

Specifically, node proximity can be calculated based on node embedding or empirically calculated based on observed edges. Minimizing the differences between the two types of proximities preserves the corresponding proximity. 

For the first-order proximity, it can be computed using node embedding as defined in Eq. \ref{eq:pfp}. 
The empirical probability is $\hat{p}^{(1)}({v_i},{v_j})= A_{ij}/\sum\nolimits_{{e_{ij}} \in E} {A_{ij}}$, where $A_{ij}$ is the weight of edge $e_{ij}$. The smaller the distance between $p^{(1)}$ and $\hat{p}^{(1)}$ is, the better first-order proximity is preserved. Adopting KL-divergence as the distance function to calculate the differences between $p^{(1)}$ and $\hat{p}^{(1)}$  and omitting some constants, the objective function to preserve the first-order proximity in graph embedding is:
\vspace{-1mm}
\begin{equation}
\label{eq:mlfo}
\textstyle
\mathcal{O}_{min}^{(1)} = \mathop {\min }- \sum\nolimits_{{e_{ij}} \in E} {A_{ij}\log {p^{(1)}}({v_i},{v_j})}. 
\end{equation} 
Similarly, the second-order proximity of $v_i$ and $v_j$ is the conditional probability of $v_j$ generated by node $v_i$ (Eq. \ref{eq:psp}).
The empirical probability of $\hat{p}^{(2)}({v_i}|{v_j})$ is calculated as $\hat{p}^{(2)}({v_j}|{v_i})= A_{ij}/d_i$, where $d_i=\sum \nolimits_{e_{ik} \in E}A_{ik}$ is the out-degree (or degree in the case of undirected graph) of node $v_i$. Similar to Eq. \ref{eq:softmax}, it is very expensive to calculate Eq. \ref{eq:psp} and negative sampling is again adopted for approximate computation to improve the efficiency. By minimizing the KL divergence between $p^{(2)}({v_j}|{v_i})$ and $\hat{p}^{(2)}({v_j}|{v_i})$, the objective function to preserve second-order proximity is:
\vspace{-1mm}
\begin{equation}
\textstyle
\label{eq:mlso}
\mathcal{O}_{min}^{(2)} = \mathop {\min } - \sum_{{e_{ij}} \in E} {A_{ij}\log {p^{(2)}}({v_j}|{v_i})}. 
\end{equation}

\begin{table}[t]
\caption{Edge reconstruction based graph embedding. $\mathcal{O}_{*}^{*}(\mathcal{T}_{v_i}^{v},\mathcal{T}_{v_j}^{v})$ refers to one of Eq. \ref{eq:mlfo1}, Eq. \ref{eq:mlso1} $\sim$ Eq. \ref{eq:mbrl}; \emph{e.g.}, $\mathcal{O}_{min}^{(2)}$(word-label) refers to Eq. \ref{eq:mlso} with a word node and a label node. $\mathcal{T}_{v_i}^{v}$ denote the type of node $v_i$. }
\vspace{-3mm}
\label{tab:erbge}
\centering
\tabcolsep=0.05cm
\scriptsize
\renewcommand{\arraystretch}{1.2}
\begin{tabular} {p{1.5cm}|p{5.9cm}|p{1.2cm}} \hline
\textbf{GE Algorithm}& \textbf{Objectives} & \textbf{Order of Proximity}  \\ \hline
PALE \cite{DBLP:conf/ijcai/ManSLJC16}& $\mathcal{O}_{max}^{(1)}$(node, node)& \multirow{2}{*}{$1^{st}$} \\\cline{1-2}
 NRCL \cite{Wei:2017:CVL:3038912.3052575} & $\mathcal{O}_{rank}$(node, neighbour node) +  $\mathcal{O}$(attribute loss) & \\\hline
PTE \cite{pte} &$\mathcal{O}_{min}^{(2)}$(word, word) + $\mathcal{O}_{min}^{(2)}$(word, document) + $\mathcal{O}_{min}^{(2)}$(word, label) & \\ \cline{1-2}
APP\cite{DBLP:conf/aaai/ZhouLLLG17} & $\mathcal{O}_{max}^{(2)}$(node, node)) & \\ \cline{1-2}
GraphEmbed \cite{Zhang:2017:RPA:3038912.3052601} & $\mathcal{O}_{min}^{(2)}$(word, word) + $\mathcal{O}_{min}^{(2)}$(word, time) + $\mathcal{O}_{min}^{(2)}$(word, location) + $\mathcal{O}_{min}^{(2)}$(time, location) + $\mathcal{O}_{min}^{(2)}$(location, location) + $\mathcal{O}_{min}^{(2)}$(time, time) & \multirow{11}{*}{$2^{nd}$}  \\ \cline{1-2}
\cite{DBLP:conf/huc/OchiNYSARM16,jcnss16geo} & $\mathcal{O}_{min}^{(2)}$(station, company),  $\mathcal{O}_{min}^{(2)}$(station, role),  $\mathcal{O}_{min}^{(2)}$(destination, boarding) & \\ \cline{1-2}
PLE \cite{Ren:2016:LNR:2939672.2939822} &  $\mathcal{O}_{rank}$(mention-type) + $\mathcal{O}_{min}^{(2)}$(mention, feature) + $\mathcal{O}_{min}^{(2)}$(type, type) & \\ \cline{1-2}
IONE \cite{DBLP:conf/ijcai/LiuCLL16} & $\mathcal{O}_{min}^{(2)}$(node, node) + $\mathcal{O}$(anchor align) & \\ \cline{1-2}
HEBE \cite{DBLP:conf/icdm/GuiLTJNH16} & $\mathcal{O}_{min}^{(2)}$(node, other nodes in one hyperedge) &  \\ \cline{1-2}
GAKE \cite{DBLP:conf/coling/FengHYZ16} & $\mathcal{O}_{max}^{(2)}$(node, neighbour context)+$\mathcal{O}_{max}^{(2)}$(node, path context)+$\mathcal{O}_{max}^{(2)}$(node, edge context)&\\ \cline{1-2}
CSIF \cite{DBLP:journals/ijmir/ZhaoZWHC16} & $\mathcal{O}_{max}^{(2)}$(user pair, diffused content) \\ \cline{1-2}
ESR \cite{Xiong:2017:ESR:3038912.3052558} &$\mathcal{O}_{max}^{(2)}$(entity, author) + $\mathcal{O}_{max}^{(2)}$(entity, entity) +  $\mathcal{O}_{max}^{(2)}$(entity, words) + $\mathcal{O}_{max}^{(2)}$(entity, venue) \\\hline
 line \cite{line} & $\mathcal{O}_{min}^{(1)}$(node, node) + $\mathcal{O}_{min}^{(2)}$(node, node)) & \multirow{2}{*}{$1^{st} + 2^{nd}$} \\\cline{1-2}
 EBPR \cite{DBLP:conf/sigir/ZhangW16} & $\mathcal{O}$(AUC ranking) + $\mathcal{O}_{max}^{(2)}$(node, node) + $\mathcal{O}_{max}^{(2)}$(node, node context) &\\\hline
\cite{DBLP:conf/emnlp/BordesCW14} & $\mathcal{O}_{rank}$(question, answer)& $1^{st} + 2^{nd}$ + higher \\\hline
\end{tabular}
\vspace{-2mm}
\end{table}

\subsubsection{Minimizing Margin-based Ranking Loss}
\label{sec:mlml}
In the margin-based ranking loss optimization, the edges of the input graph indicate the relevance between a node pair. Some nodes in the graph are usually associated with a set of relevant nodes.  E.g., in a cQA site, a set of answers are marked as relevant to a given question. The insight of the loss is straightforward.

\vspace{0.05in}\noindent
\textbf{Insight:} \textit{A node's embedding is more similar to the embedding of relevant nodes than that of any other irrelevant node.}

Denote $s(v_i,v_j)$ as the similarity score for node $v_i$ and $v_j$, $V_i^+$ as the set of nodes relevant to $v_i$ and $V_i^-$ as the irrelevant nodes set. The margin-based ranking loss is defined as:
\vspace{-1mm}
\begin{equation}
\textstyle
\label{eq:mbrl}
\mathcal{O}_{rank} = \min\sum\limits_{v_i^+ \in V_i^+} \sum\limits_{ v_i^- \in V_i^-} \max \{0,\gamma - s(v_i, v_i^+)+s(v_i, v_i^-)\},
\end{equation}
where $\gamma$ is the margin. Minimizing the loss rank encourages a large margin between $s(v_i,v_i^+)$ and $s(v_i,v_i^-)$, and thus enforces $v_i$ to be embedded closer to its relevant nodes than to any other irrelevant nodes.

In Table \ref{tab:erbge}, we summarize existing edge reconstruction based graph embedding methods, based on their objective functions and preserved node proximity. In general, most methods use one of the above objective functions (Eq. \ref{eq:mlfo1}, Eq. \ref{eq:mlso1} $\sim$ Eq. \ref{eq:mbrl}).  \cite{DBLP:conf/sigir/ZhangW16} optimizes an AUC ranking loss, which is a substitution loss for margin based ranking loss (Eq. \ref{eq:mbrl}). Note that when another task is simultaneously optimized during graph embedding, that task-specific objective will be incorporated in the overall objective. For instance, \cite{DBLP:conf/ijcai/LiuCLL16} aims to align two graphs. Hence an network alignment objective function is optimized together with $\mathcal{O}_{min}^{(2)}$ (Eq. \ref{eq:mlso}). 

It is worth noting that most of the existing knowledge graph embedding methods choose to optimize margin-based ranking loss. Recall that a knowledge graph $\mathcal{G}$ consists of a set of triplets $<h,r,t>$ denoting the head entity $h$ is linked to a tail entity $t$ by a relation $r$. Embedding $\mathcal{G}$ can be interpreted as preserving the ranking of a true triplet $<h,r,t>$ over a false triplet $<h',r,t'>$ that does not exist in $\mathcal{G}$. 
Particularly, in knowledge graph embedding, similar to $s(v_i, v_i^+)$ in Eq. \ref{eq:mbrl}, an energy function is designed for a triplet $<h,r,t>$ as $f_r(h,t)$. There is a slight difference between these two functions. $s(v_i, v_i^+)$ denotes the \textbf{similarity} score between the embedding of node $v_i$ and $v_i^+$, while $f_r(h,t)$ is the \textbf{distance} score of the embedding of $h$ and $t$ in terms of relation $r$. One example of $f_r(h,t)$ is ${\| {h + r - t} \|_{l1}}$, where relationships are represented as \textit{translations in the embedded space} \cite{DBLP:conf/nips/BordesUGWY13}. \eat{I.e., a triplet $<h,r,t>$ indicates that the embedding of $t$ should be close to the embedding of $h$ plus the embedding of $r$.} Other options of $f_r(h,t)$ are summarized in Table \ref{tab:kgemrl}. Consequently, for knowledge graph embedding, Eq. \ref{eq:mbrl} becomes:
\vspace{-1mm}
\begin{equation}
\textstyle
\label{eq:kge1}
%\mathcal{O}_{rank}^{kg} = \min\sum\limits_{<h,r,t> \in \mathcal{S}} \sum\limits_{ <h',r,t'> \notin \mathcal{S}} \max \{0,\gamma + f_r(h,t)-f_r(h',t')\}
\mathcal{O}_{rank}^{kg} = \min\sum\limits_{\substack{<h,r,t> \in \mathcal{S}, \\ <h',r,t'> \notin \mathcal{S}}} \max \{0,\gamma + f_r(h,t)-f_r(h',t')\}, 
\end{equation}
where $\mathcal{S}$ is the triples in the input knowledge graph. Existing knowledge graph embedding methods mainly optimize Eq. \ref{eq:kge1} in their work. The difference among them is how they define $f_r(h,t)$ as summarized in Table \ref{tab:kgemrl}. More details about the related work in knowledge graph embedding has been thoroughly reviewed in \cite{DBLP:journals/tkde/WangMWG17}.

\begin{table}[t]
\caption{Knowledge graph embedding using margin-based ranking loss.}
\vspace{-3mm}
\label{tab:kgemrl}
\centering
\tabcolsep=0.05cm
\scriptsize
\renewcommand{\arraystretch}{1.2}
\begin{tabular} {  l|c } \hline
\textbf{GE Algorithm}& \textbf{Energy Function} $\mathbf{{f_r}(h,t)}$ \\ \hline
TransE \cite{DBLP:conf/nips/BordesUGWY13} &${\| {h + r - t} \|_{l1}}$ \\\hline
TKRL \cite{DBLP:conf/ijcai/XieLS16} &$\| {{M_{rh}}h + r - {M_{rt}}t} \|$\\ \hline
TransR \cite{DBLP:conf/aaai/LinLSLZ15} & $ \| {h{M_r} + r - t{M_r}} \|_2^2$\\\hline
CTransR \cite{DBLP:conf/aaai/LinLSLZ15} & $ \| {h{M_r} + {r_c} - t{M_r}} \|_2^2 + \alpha \| {{r_c} - r} \|_2^2$ \\\hline
TransH \cite{DBLP:conf/aaai/WangZFC14} & $ \| {(h - w_r^Th{w_r}) + {d_r} - (t - w_r^Tt{w_r})} \|_2^2$ \\\hline
SePLi \cite{DBLP:conf/aaai/WuSYLZZ15} & $ \frac{1}{2}{\| {{W_i}{e_{ih}} + {b_i} - {e_{it}}} \|^2}$ \\\hline
TransD \cite{DBLP:conf/acl/JiHXL015} & $\| {{M_{rh}}h + r - {M_{rt}}t} \|_2^2$ \\\hline
TranSparse \cite{DBLP:conf/aaai/JiLH016}& $\| {M_r^h(\theta _r^h)h + r - M_r^t(\theta _r^t)t} \|_{{l_1}/2}^2$\\ \hline
m-TransH \cite{DBLP:conf/ijcai/WenLMCZ16} & ${\| {\sum_{\rho  \in \mathcal{M}({R_r})} {{a_r}(\rho ){\mathbb{P}_{{n_r}}}(t(\rho )) + {b_r}} } \|^2},t \in {\mathcal{N}^{\mathcal{M}({R_r})}}$ \\ \hline
DKRL \cite{DBLP:conf/aaai/XieLJLS16} & $\|h_d+r-t_d \|+\|h_d+r-t_s \|+\|h_s+r-t_d \|$ \\\hline
ManifoldE \cite{DBLP:conf/ijcai/0005HZ16}  & Sphere: $ \| \varphi(h) + \varphi(r) - \varphi(t)  \|^2$ \\
 & Hyperplane: $(\varphi(h)+\varphi(r_{head}))^T(\varphi(t)+\varphi(r_{tail}))$ \\ 
& $\varphi$ is the mapping function to Hilbert space \\ \hline
TransA \cite{DBLP:conf/aaai/JiaWLJC16} &${\| {h + r - t} \|}$ \\\hline
puTransE \cite{rlfmerri} & ${\| {h + r - t} \|}$ \\\hline
KGE-LDA \cite{DBLP:conf/aaai/YaoZWJZZC17} & ${\| {h + r - t} \|_{l1}}$\\\hline
SE \cite{DBLP:conf/aaai/BordesWCB11} & $ {\| {{R_u}h - {R_u}t} \|_{l1}}$\\ \hline
SME \cite{DBLP:journals/ml/BordesGWB14} linear & ${({W_{u1}}r + {W_{u2}}h + {b_u})^T}({W_{v1}}r + {W_{v2}}t + {b_v})$ \\
SME \cite{DBLP:journals/ml/BordesGWB14} bilinear&  ${({W_{u1}}r + {W_{u2}}h + {b_u})^T}({W_{v1}}r + {W_{v2}}t + {b_v})$ \\ \hline
SSP \cite{DBLP:conf/aaai/0005HMZ17} &$  - \lambda \| {e - {s^T}es} \|_2^2 + \| e \|_2^2$, $S({s_h},{s_t}) = \frac{{{s_h} + {s_t}}}{{\| {{s_h} + {s_t}} \|_2^2}}$\\\hline
NTN \cite{DBLP:conf/nips/SocherCMN13} & $u_r^T\tanh ({h^T}{W_r}t + {W_{rh}}h + {W_{rt}}t + {b_r})$ \\\hline
HOLE \cite{Nickel:2016:HEK:3016100.3016172} & ${r^T}(h \star t)$, where $\star$ is circular correlation\\\hline
MTransE \cite{ijcai2017-209}& ${\| {h + r - t} \|_{l1}}$ \\\hline
\end{tabular}
\vspace{-1mm}
\end{table}
 
Note that some studies jointly optimize the ranking loss (Eq. \ref{eq:kge1}) and other objectives to preserve more information. E.g., SSP \cite{DBLP:conf/aaai/0005HMZ17} optimizes a topic model loss together with Eq. \ref{eq:kge1} to utilize textual node descriptions for embedding. \cite{ijcai2017-209} categorizes monolingual relations and uses linear transformation to learn cross-lingual alignment for entities and relations. There also exists some work which defines a matching degree score rather than an energy function for a triplet $<h,r,t>$. E.g., \cite{pmlr-v70-liu17d} defines a bilinear score function $v_h^T{W_r}{v_t}$ It adds a normality constraint and a commutativity constraint to impose analogical structures among the embedding. ComplEx \cite{Trouillon:2016:CES:3045390.3045609} extends embedding to complex domain and defines the real part of $v_h^T{W_r}{v_t}$ as the score.

\vspace{0.05in}\noindent
\textbf{Summary:} Edge reconstruction based optimization is applicable for most graph embedding settings. As far as can be observed, only graph constructed from non-relational data (Sec. \ref{sec:cg}) and whole-graph embedding (Sec. \ref{sec:wge}) have not been tried. The reason is that reconstructing manually constructed edges is not as intuitive as in other graphs. Moreover, as this technique focuses on the directly observed local edges, it is not suitable for whole-graph embedding.

\subsection{Graph Kernel}
\label{sec:gk}
\vspace{0.05in}\noindent
\textbf{Insight:} \textit{The whole graph structure can be represented as a vector containing the counts of elementary substructures that are decomposed from it.}

Graph kernel is an instance of R-convolution kernels \cite{haussler99convolution}, which is a generic way of defining kernels on discrete compound objects by recursively decomposing structured objects into ``atomic'' substructures and comparing all pairs of them \cite{dgk}. The graph kernel represents each graph as a vector, and two graphs are compared using an inner product of the two vectors. 
There are generally three types of ``atomic'' substructures defined in graph kernel.

\vspace{0.05in}\noindent
\textbf{Graphlet}.
A graphlet is an induced and non-isomorphic subgraph of size-$k$ \cite{dgk}. Suppose graph $\mathcal{G}$ is decomposed into a set of graphlets $\{G_1, G_2, \cdots, G_d \}$. Then $\mathcal{G}$ is embedded as a $d$-dimensional vector (denoted as $y_{\mathcal{G}}$) of normalized counts. The $i$-th dimension of $y_{\mathcal{G}}$ is the frequency of the graphlet $G_i$ occurring in $\mathcal{G}$.

\vspace{0.05in}\noindent
\textbf{Subtree Patterns}. In this kernel, a graph is decomposed as its subtree patterns. One example is Weisfeiler-Lehman subtree \cite{journals/jmlr/ShervashidzeSLMB11}. In particular, a relabelling iteration process is conducted on a labelled graph (i.e., a graph with discrete node labels). At each iteration, a multiset label is generated based on the label of a node and its neighbours. The new generated multiset label is a compressed label which denotes the subtree patterns, which is then used for the next iteration. Based on Weisfeiler-Lehman
test of graph isomorphism, counting the occurrence of labels in a graph is equivalent to counting the corresponding subtree structures. Suppose $h$ iterations of relabelling are performed on graph $\mathcal{G}$. Its embedding $y_{\mathcal{G}}$ contains $h$ blocks. The $i$-th dimension in the $j$-th block of $y_{\mathcal{G}}$ is the frequency of the $i$-th label being assigned to a node in the $j$-th iteration.

\vspace{0.05in}\noindent
\textbf{Random Walks}.
In the third type of graph kernels, a graph is decomposed into random walks or paths and represented as the counts of occurrence of random walks \cite{DBLP:journals/jmlr/VishwanathanSKB10} or paths \cite{DBLP:conf/icdm/BorgwardtK05} in it. Take paths as an example, suppose graph $\mathcal{G}$ is decomposed into $d$ shortest paths. Denote the $i$-th path as a triplet $<l_i^s, l_i^e, n_i>$, where $l_i^s$ and $l_i^e$ are the labels of the starting and ending nodes, $n_i$ is the length of the path. Then $\mathcal{G}$ is represented as a $d$-dimensional vector $y_{\mathcal{G}}$ where the $i$-th dimension is the frequency of the $i$-th triplet occurring in $\mathcal{G}$.

\eat{Given the graph embedding obtained from one of the three kernel methods introduced above, the graph kernel between graph $\mathcal{G}$ and $\mathcal{G}'$ is defined as:
\begin{equation}
\textstyle
\mathcal{K}(\mathcal{G}, \mathcal{G}') = y_{\mathcal{G}} \cdot y_{\mathcal{G}'}
\end{equation}
where $\cdot$ is the Euclidean dot product.}

\eat{Recently, some studies start to utilize node embedding to compare graphs. For example, \cite{DBLP:conf/aaai/NikolentzosMV17} first embeds a graph as a set of node embeddings and then uses Earth Mover’s Distance metric to calculate the similarities between two sets of vectors for graph comparison. \cite{dgk} decomposes each graph into a set of sub-structures. The list of substructures is treated as a sequence and Skipgram is applied to embed the substructure as a vector. The graphs can then be compared via the substructure similarities. \cite{DBLP:conf/kdd/JohanssonD15} utilizes existing graph embedding methods such as graph Laplacian (Sec. \ref{sec:gle}) to calculate a weight matrix, in which each element is the Euclidean distance between a pair of nodes from two graphs. After that, they use maximum weight matching of the matrix as the graph kernel to train a graph classifier. \cite{DBLP:conf/icml/DaiDS16} embeds graphs by performing a sequence of function mappings in a way similar to graphical model inference, such as mean field and belief propagation. It outputs a set of node embedding for each graph and train a regression/classification objective function to predict the continuous/discrete graph label.}

\vspace{0.05in}\noindent
\textbf{Summary:} A graph kernel is designed for whole-graph embedding (Sec. \ref{sec:wge}) only as it captures the global property of a whole graph. The type of input graph is usually a homogeneous graph (Sec. \ref{sec:hog}) \cite{dgk} or a graph with auxiliary information (Sec. \ref{sec:gwai}) \cite{journals/jmlr/ShervashidzeSLMB11}.

\subsection{Generative Model}
\label{sec:gm}
A generative model can be defined by specifying the joint distribution of the input features and the class labels, conditioned on a set of parameters \cite{bishop2007}. An example is Latent Dirichlet Allocation (LDA), in which a document is interpreted as a distribution over topics, and a topic is a distribution over words \cite{DBLP:conf/nips/BleiNJ01}.  
There are the following two ways to adopt a generative model for graph embedding.

\subsubsection{Embed Graph Into The Latent Semantic Space}

\vspace{0.05in}\noindent
\textbf{Insight:} \textit{Nodes are embedded into a latent semantic space where the distances among nodes explain the observed graph structure.}

The first type of generative model based graph embedding methods directly embeds a graph in the latent space. Each node is represented as a vector of the latent variables. In other words, it views the observed graph as generated by a model. E.g., in LDA, documents are embedded in a ``topic'' space where documents with similar words have similar topic vector representations. \cite{DBLP:conf/icdm/AlharbiZ16} designs a LDA-like model to embed a location-based social network (LBSN) graph. Specifically, the input is locations (documents), each of which contains a set of users (words) who visited that location. Users visit the same location (words appearing in the same document) due to some activities (topics). Then a model is designed to represent a location as a distribution over activities, where each activity has an attractiveness distribution over users. Consequently, both user and location are represented as a vector in the ``activity'' space. 

\subsubsection{Incorporate Latent Semantics for Graph Embedding}
\vspace{0.05in}\noindent
\textbf{Insight:} \textit{Nodes which are close in the graph and having similar semantics should be embedded closer. The node semantics can be detected from node descriptions via a generative model.}

In this line of methods, latent semantics are used to leverage auxiliary node information for graph embedding. The embedding is decided not only by the graph structure information but also by the latent semantics discovered from other sources of node information. 
For example, \cite{DBLP:conf/icdm/LeL14} proposes a unified framework which jointly integrates topic modelling and graph embedding. Its principle is that if two nodes are close in the embedded space, they will also share similar topic distribution. A mapping function from the embedded space to the topic semantic space is designed so as to correlate the two spaces.  \cite{DBLP:conf/acl/0005HZ16} proposes a generative model (Bayesian non-parametric infinite mixture embedding model) to address the issue of multiple relation semantics in knowledge graph embedding. It discovers the latent semantics of a relation and leverages a mixture of relation components for embedding. \cite{DBLP:conf/aaai/0005HMZ17} embeds a knowledge graph from both the knowledge graph triplets and the textual descriptions of entities and relations. It learns the semantic representation of text using topic modelling and restricts the triplet embedding in the semantic subspace.    

The difference between the above two directions of methods is that the embedded space is the latent space in the first way. In contrast, in the second way, the latent space is used to integrate information from different sources, and help to embed a graph to another space.

\vspace{0.05in}\noindent
\textbf{Summary:} Generative model can be used for both node embedding (Sec. \ref{sec:ne}) \cite{DBLP:conf/icdm/AlharbiZ16} and edge embedding (Sec. \ref{sec:ee}) \cite{DBLP:conf/acl/0005HZ16}. As it considers node semantics, the input graph is usually a heterogeneous graph (Sec. \ref{sec:heg}
) \cite{DBLP:conf/icdm/AlharbiZ16} or a graph with auxiliary information (Sec. \ref{sec:gwai}) \cite{DBLP:conf/aaai/0005HMZ17}.

\subsection{Hybrid Techniques and Others}
\begin{table*}[!htbp]
\caption{Comparison of graph embedding techniques.}
\vspace{-3mm}
\label{tab:get}
\centering
\tabcolsep=0.05cm
\scriptsize
\renewcommand{\arraystretch}{1.2}
\begin{tabular} {  l|l|l|l } \myhline
\textbf{Category}& \textbf{Subcategory} & \textbf{Advantages} & \textbf{Disadvantages}\\  \myhline
\multirow{ 2}{*}{matrix factorization} & graph Laplacian eigenaps& consider global node proximity & large time and space consumption\\\cline{2-2}
 &node proximity matrix factorization & \\\myhline
\multirow{ 3}{*}{deep learning} & \multirow{2}{*}{with random walk}& effective and robust, & a) only consider local context within a path \\ 
& & no feature engineering& b) hard to find optimal sampling strategy  \\ \cline{2-2} \cline{4-4}
& without random walk & & high computation cost \\ \myhline
\multirow{ 3}{*}{edge reconstruction }  & maximize edge reconstruct probability&& optimization using only observed local   \\\cline{2-2}
&minimize distance-based loss & relatively efficient training& information, i.e., edges ($1$-hop neighbour)  \\ \cline{2-2} 
& minimize margin-based ranking loss & & or ranked node pairs\\\myhline
\multirow{ 3}{*}{graph kernel} & based on graphlet& efficient, only counting the desired atom &a) substructures are not independent\\ \cline{2-2}
& based on subtree patterns & substructure    & b) embedding dimensionality grows \\\cline{2-2}
& based on random walks && ~~~~exponentially\\\myhline
\multirow{3}{*}{generative model} & embed graph in the latent space& interpretable embedding & a) hard to justify the choice of distribution \\ \cline{2-3}
& incorporate latent semantics for graph embedding&naturally leverage multiple information sources & b) require a large amount of training data \\\myhline
\end{tabular}
\vspace{-1mm}
\end{table*}
Sometimes multiple techniques are combined in one study. For example, \cite{Wei:2017:CVL:3038912.3052575} learns edge-based embedding via minimizing the margin-based ranking loss (Sec. \ref{sec:ml}), and learns attribute-based embedding by matrix factorization (Sect. \ref{sec:mf}). \cite{DBLP:conf/acl/GuoWWWG15} optimizes a margin-based ranking loss (Sec. \ref{sec:ml}) with matrix factorization based loss (Sec. \ref{sec:mf}) as regularization terms. 
\cite{DBLP:conf/aaai/ZhaoLZCHZ17} uses LSTM (Sec. \ref{sec:dl}) to learn sentences embedding in cQAs and a margin-based ranking loss (Sec. \ref{sec:ml}) to incorporate friendship relations. 
\cite{cdkge} adopts CBOW/SkipGram (Sec. \ref{sec:dl}) for knowledge graph entity embedding, and then fine-tune the embedding by minimizing a margin-based ranking loss (Sec. \ref{sec:ml}). 
\cite{Wang:2016:TRL:3060621.3060801} use word2vec (Sec. \ref{sec:dl}) to embed the textual context and TransH (Sec. \ref{sec:ml}) to embed the entity/relations so that the rich context information is utilized in knowledge graph embedding.
\cite{DBLP:conf/kdd/ZhangYLXM16} leverages the heterogeneous information in a knowledge base to improve recommendation performance. It uses TransR (Sec. \ref{sec:ml}) for network embedding, and uses auto-encoders for textual and visual embedding (Sec. \ref{sec:dl}). Finally, a generative framework (Sec. \ref{sec:gm}) is proposed to integrate collaborative filtering with items’ semantic representations. 

Apart from the introduced five categories of techniques, there exist other approaches. \cite{DBLP:journals/pr/MousaviSMB17} presents embedding of a graph by its distances to prototype graphs. \cite{DBLP:journals/pvldb/ZhaoCSZZ13} first embeds a few landmark nodes using their pairwise shortest path distances. Then other nodes are embedded so that their distances to a subset of landmarks are as close as possible to the real shortest paths. \cite{Wei:2017:CVL:3038912.3052575} jointly optimizes a link-based loss (maximizing the likelihood of observing a node's neighbours instead of its non-neighbours) and an attribute-based loss (learning a linear projection based on link-based representation). 
KR-EAR \cite{DBLP:conf/ijcai/LinLS16} distinguishes the relations in a knowledge graph as attribute-based and relation-based. It constructs a relational triple encoder (TransE, TransR) to embed the correlations between entities and relations, and an attributional triple encoder to embed the correlations between entities and attributes. Struct2vec \cite{Ribeiro:2017:SLN:3097983.3098061} considers the structral identify of nodes by a hierarchical metric for node embedding. \cite{ijcai2017-544} provides a fast embedding approach by approximating the higher-order proximity matrices.

\subsection{Summary}
We now summarize and compare all the five categories of introduced graph embedding techniques in Table \ref{tab:get} in terms of their advantages and disadvantages.

Matrix factorization based graph embedding learns the representations based on the statistics of global pairwise similarities. Hence it can outperform deep learning based graph embedding (random walk involved) in certain tasks as the latter relies on separate local context windows \cite{Bullinaria2012, DBLP:journals/tacl/LevyGD15}. However, either the proximity matrix construction or the eigendecomposition of the matrix is time and space consuming \cite{DBLP:journals/nm/DemmelDH07}, making matrix factorization inefficient and unscalable for large graphs.

Deep Learning (DL) has shown promising results among different graph embedding methods. We consider DL as suitable for graph embedding, because of its capability of automatically identifying useful representations from complex graph structures. For example, DL with random walk  (e.g., DeepWalk \cite{deepwalk}, node2vec \cite{node2vec}, metapath2vec  \cite{dong2017metapath2vec}) can automatically exploit the neighbourhood structure through sampled paths on the graph. DL without random walk can model variable-sized subgraph structures in homogeneous graphs (e.g., GCN \cite{DBLP:journals/corr/KipfW16}, struc2vec \cite{Ribeiro:2017:SLN:3097983.3098061}, GraphSAGE \cite{DBLP:conf/nips/HamiltonYL17}), or rich interactions among flexible-typed nodes in heterogeneous graphs (e.g., HNE \cite{hneda}, TransE \cite{DBLP:conf/nips/BordesUGWY13}, ProxEmbed \cite{DBLP:conf/aaai/LiuZZZCWY17}), as useful representations. On the other hand, DL also has its limitations. For DL with random walks, it typically considers a node’s local neighbours within the same path and thus overlooks the global structure information. Moreover, it is difficult to find an `optimal sampling strategy as the embedding and path sampling are not jointly optimized in a unified framework. For DL without random walks, the computational cost is usually high. The traditional deep learning architectures assume the input data on 1D or 2D grid to take advantage of GPU \cite{DBLP:journals/corr/BronsteinBLSV16}. However, graphs do not have such a grid structure, and thus require different solutions to improve the efficiency \cite{DBLP:journals/corr/BronsteinBLSV16}.

Edge reconstruction based graph embedding optimizes an objective function based on the observed edges or ranking triplets. It is more efficient compared to the previous two categories of graph embedding. However, this line of methods are trained using the directly observed local information, and thus the obtained embedding lacks the awareness of the global graph structure.

Graph kernel based graph embedding converts a graph into one single vector to facilitate the graph level analytic tasks such as graph classification. It is more efficient than other categories of techniques as it only needs to enumerate the desired atomic substructures in a graph. However, such ``bag-of-structure'' based methods have two limitations \cite{dgk}. Firstly, the substructures are not independent. For example, the graphlet of size $k+1$ can be derived from size $k$ graphlet by adding a new node and some edges. This means there exist redundant information in the graph representation. Secondly, the embedding dimensionality usually grows exponentially when the size of the substructure grows, leading to a sparse problem in the embedding.

Generative model based graph embedding can naturally leverage information from different sources (e.g., graph structure, node attribute) in a unified model. Directly embedding graphs into the latent semantic space generates the embedding that can be interpreted using the semantics. But the assumption of modelling the observation using certain distributions is hard to justify. Moreover, the generative method needs a large amount of training data to estimate a proper model which fits the data. Hence it may not work well for small graphs or a small number of graphs.

\eat{\begin{table}[!htbp]
\caption{Enumeration of Existing Graph Embedding Work in Terms of Embedding Techniques and Problem Setting}
\label{tab:etps}
\centering
\tabcolsep=0.05cm
\begin{small}
\renewcommand{\arraystretch}{1.2}
\resizebox{0.8\linewidth}{!}{%
\begin{tabular} {  |l|l|c|c|c|c|c|c|c|c| } \hline
& & \multicolumn{8}{|c|}{Problem Setting} \\ \cline{3-10}
&&  3.1 & 3.2 & 3.3 &3.4 & 3.5 & 3.6 & 3.7 & 3.8  \\ \hline
\multirow{5}{*}{\rotatebox[origin=c]{90}{Techniques}}&4.1 &\checkmark & & &\checkmark &\checkmark & &\checkmark & \\\cline{2-10}
&4.2 & \checkmark &\checkmark &\checkmark & &\checkmark &\checkmark &\checkmark &\checkmark \\\cline{2-10}
&4.3 & \checkmark &\checkmark &\checkmark & &\checkmark &\checkmark &\checkmark &\\\cline{2-10}
&4.4 &\checkmark & &\checkmark & & & & &\checkmark \\\cline{2-10}
&4.5 & &\checkmark &\checkmark & &\checkmark &\checkmark & &\\\hline
\end{tabular}}
\end{small}
\end{table}}

\section{Applications}
\label{sec:apps}
Graph embedding benefits a wide variety of graph analytics applications as the vector representations can be processed efficiently in both time and space. In this section, we categorize the graph embedding enabled applications as node related, edge related and graph related.

\subsection{Node Related Applications}

\subsubsection{Node Classification}
Node classification is to assign a class label to each node in a graph based on the rules learnt from the labelled nodes. Intuitively, ``similar'' nodes have the same labels. It is one of the most common applications discussed in graph embedding literatures. In general, each node is embedded as a low-dimensional vector. Node classification is conducted by applying a classifier on the set of labelled node embedding for training. The example classifiers include SVM (\cite{DBLP:conf/aaai/WangCWP0Y17,hneda,nrlrti,sdne,DBLP:conf/ijcai/PanWZZW16,DBLP:journals/tomccap/ZhangSLWC16,DBLP:conf/icdm/GuiLTJNH16,DBLP:conf/huc/OchiNYSARM16,jcnss16geo,DBLP:conf/icdm/ZhangYZZ16,DBLP:conf/ijcai/HanS16,DBLP:journals/tkde/ChenTTJ15,DBLP:conf/kdd/SunJY08,DBLP:conf/aaai/YaoZWJZZC17}), logistic regression (\cite{deepwalk,line,pte,DBLP:conf/aaai/WangCWP0Y17,grarep,node2vec,sdne,tpavnz,DBLP:conf/icdm/GuiLTJNH16,DBLP:conf/ijcnn/JinLLZZW16,DBLP:conf/aaai/0005HMZ17}) and k-nearest neighbour classification (\cite{DBLP:conf/icdm/LeL14,DBLP:journals/pami/WilsonHPD14}). Then given the embedding of an unlabelled node, the trained classifier can predict its class label.
In contrast to the above sequential processing of first node embedding then node classification, some other work ( \cite{rsslge,DBLP:journals/corr/KipfW16,DBLP:conf/acl/LiZZ16,DBLP:conf/ijcai/TuZLS16,DBLP:journals/corr/MontiBMRSB16}) designs a unified framework to jointly optimize graph embedding and node classification, which learns a classification-specific representation for each node.

\subsubsection{Node Clustering}
Node clustering aims to group similar nodes together, so that nodes in the same group are more similar to each other than those in other groups. As an unsupervised algorithm, it is applicable when the node labels are unavailable. After representing nodes as vectors, the traditional clustering algorithms can then be applied on the node embedding. Most existing work \cite{DBLP:conf/aaai/WangCWP0Y17,DBLP:conf/aaai/NieZL17,grarep,hneda,dnnflgr,DBLP:conf/aaai/TianGCCL14,DBLP:journals/tkde/ChenTTJ15} adopts k-means as the clustering algorithm. In contrast, \cite{Wei:2017:CVL:3038912.3052575} and \cite{DBLP:conf/mm/TangNJ16} jointly optimize clustering and graph embedding in one objective to learn a clustering-specific node representation.

\subsubsection{Node Recommendation/Retrieval/Ranking}
The task of node recommendation is to recommend top K nodes of interest to a given node based on certain criteria such as similarity \cite{atpge,DBLP:conf/aaai/ZhouLLLG17,DBLP:conf/acl/LiZZ16,DBLP:journals/pvldb/ZhaoCSZZ13,rlfmerri,DBLP:conf/icdm/GuiLTJNH16}. 
In real-world scenarios, there are various types of recommended node, such as research interests for researchers \cite{mmbeflskg}, items for customers \cite{DBLP:conf/aaai/ZhouLLLG17,DBLP:conf/sigir/ZhangW16}, images for curation network users \cite{DBLP:conf/iccv/GengZBC15}, friends for social network users \cite{DBLP:conf/aaai/ZhouLLLG17}, and documents for a query \cite{Xiong:2017:ESR:3038912.3052558}. It is also popular in community-based question answering. Given a question, they predict the relative rank of users (\cite{DBLP:conf/aaai/LuK17,DBLP:conf/ijcai/ZhaoYCHZ16}) or answers (\cite{DBLP:conf/aaai/ZhaoLZCHZ17,cqahsnl}). In proximity search \cite{DBLP:conf/aaai/LiuZZZCWY17,DBLP:conf/aaai/WuSYLZZ15}, they rank the nodes of a particular type (e.g., ``user'') for a given query node (e.g., ``Bob'') and a proximity class (e.g., ``schoolmate''), e.g., ranking users who are the schoolmates of Bob. And there is some work focusing on cross-modal retrieval \cite{hneda,DBLP:journals/tip/WuLSYZRZ16,DBLP:conf/mm/CaiHH07,DBLP:journals/tomccap/ZhangSLWC16}, e.g., keyword-based image/video search. 

A specific application which is popularly discussed in knowledge graph embedding is entity ranking \cite{DBLP:conf/ijcai/XieLS16,DBLP:conf/acl/GuoWWWG15,DBLP:conf/aaai/0005HMZ17,DBLP:journals/tkde/GuoWWWG17,Wang:2016:TRL:3060621.3060801}
\eat{\cite{cdkge,DBLP:conf/aaai/BordesWCB11,DBLP:conf/nips/BordesUGWY13,DBLP:journals/ml/BordesGWB14,DBLP:conf/acl/0005HZ16,DBLP:conf/ijcai/XieLS16,DBLP:conf/aaai/LinLSLZ15,DBLP:conf/acl/GuoWWWG15,DBLP:conf/aaai/0005HMZ17,DBLP:journals/tkde/GuoWWWG17,DBLP:conf/aaai/WangZFC14,DBLP:conf/coling/FengHYZ16,DBLP:conf/aaai/WuSYLZZ15,DBLP:conf/aaai/ShiW17,Wang:2016:TRL:3060621.3060801}}. Recall that a knowledge graph consists of a set of triplets $<h,r,t>$. Entity ranking aims to rank the correct missing entities given the other two components in a triplet higher than the false entities. E.g., it returns the true $h$'s among all the candidate entities given $r$ and $t$, or returns the true $t$'s given $r$ and $h$.

\subsection{Edge Related Applications}
\eat{The above applications are all related to individual nodes.} Next we introduce edge related applications in which an edge or a node pair is involved.
\subsubsection{Link Prediction}
Graph embedding aims to represent a graph with low-dimensional vectors, but interestingly its output vectors can also help infer the graph structure. In practice, graphs are often incomplete; e.g., in social networks, friendship links can be missing between two users who actually know each other. In graph embedding, the low-dimensional vectors are expected to preserve different orders of network proximity (e.g., DeepWalk \cite{deepwalk}, LINE \cite{line}), as well as different scales of structural similarity (e.g., GCN \cite{DBLP:journals/corr/KipfW16}, struc2vec \cite{Ribeiro:2017:SLN:3097983.3098061}). Hence, these vectors encode rich information about the network structure, and they can be used to predict missing links in the incomplete graph. Most attempts on graph embedding driven link prediction are on homogeneous graphs \cite{DBLP:conf/aaai/ZhouLLLG17,DBLP:journals/pvldb/ZhaoCSZZ13,node2vec,tpavnz}. For example, \cite{node2vec} predicts the friendship relation between two users. Relatively fewer graph embedding work deals with heterogeneous graph link prediction. For example, on a heterogeneous social graph, ProxEmbed \cite{DBLP:conf/aaai/LiuZZZCWY17} tries to predict the missing links of certain semantic types (e.g., schoolmates) between two users, based on the embedding of their connecting paths on the graph. D2AGE \cite{vincentaaai18}  solves the same problem by embedding two users’ connecting directed acyclic graph structure.

\subsubsection{Triple Classification}
Triplet classification \cite{cdkge,DBLP:conf/ijcai/XieLS16,DBLP:conf/aaai/LinLSLZ15,DBLP:conf/acl/GuoWWWG15,DBLP:journals/tkde/GuoWWWG17,DBLP:conf/aaai/WangZFC14,DBLP:conf/coling/FengHYZ16,Wang:2016:TRL:3060621.3060801} is a specific application for knowledge graph. It aims to classify whether an unseen triplet $<h,r,t>$ is correct or not, i.e., whether the relation between $h$ and $t$ is $r$.

\subsection{Graph Related Applications}

\subsubsection{Graph Classification}
Graph classification assigns a class label to a whole graph. This is important when the graph is the unit of data. For example, in \cite{DBLP:conf/aaai/NikolentzosMV17}, each graph is a chemical compound, an organic molecule or a protein structure. 
In most cases, whole-graph embedding is applied to calculate graph level similarity \cite{dgk,lcnnfg,DBLP:conf/icml/DaiDS16,journals/jmlr/ShervashidzeSLMB11,DBLP:journals/pr/MousaviSMB17}. Recently, some work starts to match node embedding for graph similarity \cite{DBLP:conf/kdd/JohanssonD15,DBLP:conf/aaai/NikolentzosMV17}. Each graph is represented as a set of node embedding vectors. Graphs are compared based on two sets of node embedding. \cite{dgk} decomposes a graph into a set of sub-structures and then embed each substructure as a vector and compare graphs via substructure similarities.

\subsubsection{Visualization}
Graph visualization generates visualizations of a graph on a low dimensional space \cite{sdne,dnnflgr,lcnnfg,DBLP:conf/ijcai/TuZLS16,DBLP:conf/ijcai/PanWZZW16,DBLP:conf/icdm/LeL14}. Usually, for visualization purpose, all nodes are embedded as 2D vectors and then plotted in a 2D space with different colours indicating nodes' categories. It provides a vivid demonstration of whether nodes belonging to the same category are embedded closer to each other.

\subsection{Other Applications}
Above are some general applications that are commonly discussed in existing work. Depending on the information carried in the input graph, more specific applications may exist. Below are some example scenarios.

\textbf{Knowledge graph related}: \cite{DBLP:conf/aaai/LinLSLZ15} and \cite{DBLP:conf/aaai/WangZFC14} extract relational fact from large-scale plain text. \cite{rsslge} extracts medical entities from text. \cite{Xiong:2017:ESR:3038912.3052558} links natural language text with entities in a knowledge graph. \cite{DBLP:journals/ml/BordesGWB14} focuses on de-duplicating entities that are equivalent in a knowledge graph. \cite{Ren:2016:LNR:2939672.2939822} jointly embeds entity mentions, text and entity types to estimate the true type-path for each mention from its noisy candidate type set. E.g., the candidate types for ``Trump'' are $\{$person, politician, businessman, artist, actor$\}$. For the mention ``Trump'' in sentence ``Republican presidential candidate Donald Trump spoke during a campaign event in Rock Hill.'', only $\{$person, politician$\}$ are correct types.

\textbf{Multimedia network related}: \cite{Zhang:2017:RPA:3038912.3052601} embeds the geo-tagged social media (GTSM) records $<$time, location, message$>$ which enables them to recover the missing component from a GTSM triplet given the other two. It can also classify the GTSM records, e.g., whether a check-in record is related to ``Food'' or ``Shop''. \cite{DBLP:journals/pami/YanXZZYL07} uses graph embedding to reduce data dimensionality for face recognition. \cite{Lin:2005:SML:1101149.1101193} maps images into a semantic manifold that faithfully grasps users' preferences to facilitate  content-based image retrieval. 

\textbf{Information propagation related}: \cite{DBLP:conf/www/LiMGM17} predicts the increment of a cascade size after a given time interval.  \cite{DBLP:journals/ijmir/ZhaoZWHC16} predicts the propagation user and identifies the domain expert by embedding the social interaction graph.

\textbf{Social networks alignment:} Both \cite{DBLP:conf/ijcai/LiuCLL16} and \cite{DBLP:conf/ijcai/ManSLJC16} learn node embedding to align users across different social networks, i.e., to predict whether two user accounts in two different social networks are owned by the same user.

\textbf{Image related}: Some work embeds graphs constructed from images, and then use the embedding for image classification (\cite{DBLP:journals/tkde/ChenTTJ15, DBLP:conf/nips/DefferrardBV16}), image clustering \cite{DBLP:journals/tkde/AllabLN17}, image segmentation \cite{DBLP:conf/iccv/YuFL15}, pattern recognition \cite{DBLP:journals/corr/MontiBMRSB16}, and so on.

\section{Future Directions}
\label{sec:fd}
In this section, we summarize four future directions for the field of graph embedding, including computation efficiency, problem settings, techniques and application scenarios.

\vspace{0.05in}\noindent
\textbf {Computation.}
The deep architecture, which takes the geometric input (e.g., graph), suffers the low efficiency problem. Traditional deep learning models (designed for Euclidean domains) utilize the modern GPU to optimize their efficiency by assuming that the input data are on a 1D or 2D grid. However, graphs do not have such a kind of grid structure and thus the deep architecture designed for graph embedding needs to seek alternative solutions to improve the model efficiency. 
\cite{DBLP:journals/corr/BronsteinBLSV16} suggested that the computational paradigms developed for large-scale graph processing can be adopted to facilitate efficiency improvement in deep learning models for graph embedding. 

\vspace{0.05in}\noindent
\textbf {Problem settings.}
The dynamic graph is one promising setting for graph embedding. Graphs are not always static, especially in real life scenarios, e.g., social graphs in Twitter, citation graphs in DBLP. Graphs can be dynamic in terms of graph structure or node/edge information. On the one hand, the graph structure may evolve over time, i.e., new nodes/edges appear while some old nodes/edges disappear. On the other hand, the nodes/edges may be described by some time-varying information. Existing graph embedding mainly focuses on embedding the static graph and the settings of dynamic graph embedding are overlooked.
Unlike static graph embedding, the techniques for dynamic graphs need to be scalable and better to be incremental so as to deal with the dynamic changes efficiently. This makes most of the existing graph embedding methods, which suffer from the low efficiency problem, not suitable anymore. How to design effective graph embedding methods in dynamic domains remains an open question.

\vspace{0.05in}\noindent
\textbf {Techniques.}
Structure awareness is important for edge reconstruction based graph embedding. Current edge reconstruction based graph embedding methods are mainly based on the edges only, e.g., $1$-hope neighbours in a general graph, a ranked triplet ($<h,r,t>$ in a knowledge graph, and $(v_i, v_i^{+}, v_i^{-})$ in a cQA graph. Single edges only provide local neighbourhood information to calculate the first- and second-order proximity. The global structure of a graph (e.g., paths, tree, subgraph patterns) is omitted. 
Intuitively, a substructure contains richer information than one single edge. Some work attempts to explore the path information in knowledge graph embedding (\cite{cdkge,DBLP:conf/aaai/ShiW17, DBLP:conf/coling/FengHYZ16, DBLP:conf/aaai/WuSYLZZ15}). However, most of them use deep learning models (\cite{cdkge, DBLP:conf/coling/FengHYZ16, DBLP:conf/aaai/ShiW17}) which suffer the low efficiency issue as discussed earlier. How to design the non-deep learning based methods that can take advantage of the expressive power of graph structure is a question. \cite{DBLP:conf/aaai/WuSYLZZ15} provides one example solution. It minimizes both pairwise and long-range loss to capture pairwise relations and long-range interactions between entities. Note that in addition to the list/path structure, there are various kinds of substructures which carry different structure information. For example, SPE \cite{spe_www2018} has tried to introduce a subgraph-augmented path structure for embedding the proximity between two nodes in a heterogeneous graph, and it shows better performance than embedding simple paths for semantic search tasks. In general, an efficient structure-aware graph embedding optimization solution, together with the substructure sampling strategy, is in needed. 

\vspace{0.05in}\noindent
\textbf {Applications.}
Graph embedding has been applied in many different applications. It is an effective way to learn the representations of data with consideration of their relations. 
Moreover, it can convert data instances from different sources/platforms/views into one common space so that they are directly comparable. For example, \cite{DBLP:journals/tip/WuLSYZRZ16, DBLP:journals/pvldb/ZhaoCSZZ13, DBLP:journals/tomccap/ZhangSLWC16} use graph embedding for cross-modal retrieval, such as content-based image retrieval, keyword-based image/video search. The advantages of using graph embedding for representation learning 
is that the graph manifold of the training data instances are preserved in the representations and can further benefit the follow-up applications.
Consequently, graph embedding can benefit the tasks which assume the input data instances are correlated with certain relations (i.e., connected by certain links). It is of great importance to exploring the application scenarios which benefit from graph embedding, as it provides effective solutions to the conventional problems from a different perspective.

\section{Conclusions}
\label{sec:conc}
In this survey, we conduct a comprehensive review of the literature in graph embedding. We provide a formal definition to the problem of graph embedding and introduce some basic concepts. More importantly, we propose two taxonomies of graph embedding, categorizing existing work based on problem settings and embedding techniques respectively. In terms of problem setting taxonomy, we introduce four types of embedding input and four types of embedding output and summarize the challenges faced in each setting. For embedding technique taxonomy, we introduce the work in each category and compare them in terms of their advantages and disadvantages. After that, we summarize the applications that graph embedding enables. Finally, we suggest four promising future research directions in the field of graph embedding in terms of computation efficiency, problem settings, techniques and application scenarios.

\ifCLASSOPTIONcaptionsoff
  \newpage
\fi

\bibliographystyle{IEEEtran}
\begin{small}
\bibliography{sigproc}
\end{small}

\vspace{-10mm}
\begin{IEEEbiography}[{\includegraphics[width=0.9in, height=1.25in,clip,keepaspectratio]{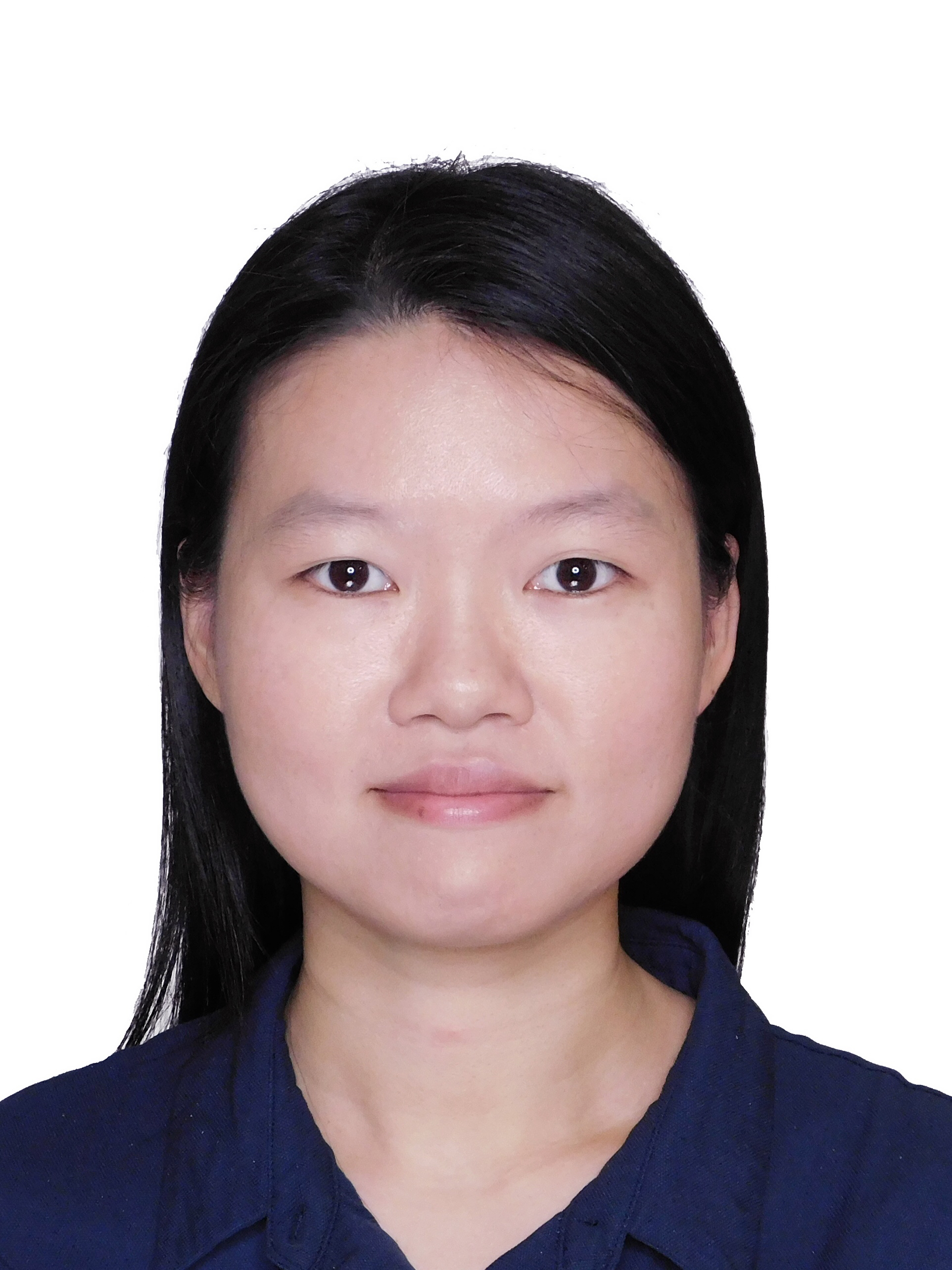}}]{Hongyun Cai} 
received the Ph.D.~degree in Computer Science from the University of
Queensland in 2016. She is currently a postdoctoral researcher at the
Advanced Digital Sciences Center, Singapore. Her research focuses on graph mining, social data management and analysis.
\end{IEEEbiography}
\vspace{-15mm}

\begin{IEEEbiography}[{\includegraphics[width=0.9in, height=1.25in,clip,keepaspectratio]{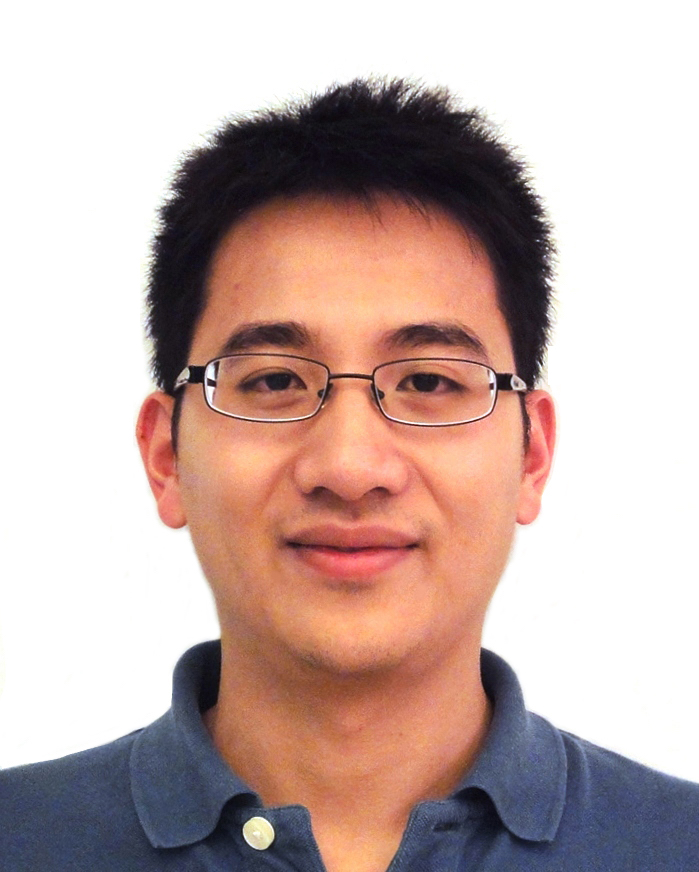}}]{Vincent
W.~Zheng} received the Ph.D.~degree in Computer Science from the Hong Kong
University of Science and Technology in 2011. He is a senior research scientist at the
Advanced Digital Sciences Center, Singapore, and a research affiliate at the University of Illinois at Urbana-Champaign.
His research interests include graph mining, information extraction, ubiquitous
computing and machine learning.
\end{IEEEbiography}
\vspace{-15mm}

\begin{IEEEbiography}[{\includegraphics[width=0.8in, height=1.25in,clip,keepaspectratio]{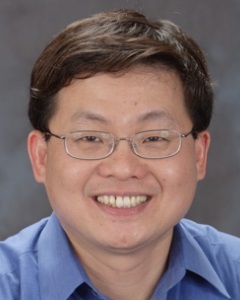}}]{Kevin Chen-Chuan Chang}
is a Professor in University of Illinois at Urbana-Champaign.
His research addresses large-scale information access, for search, mining, and
integration across structured and unstructured big data including Web data and
social media. He also co-founded Cazoodle
for deepening vertical data-aware search over the Web.
\end{IEEEbiography}

\end{document}